\documentclass[10pt,twocolumn,letterpaper]{article}

\usepackage[accsupp]{axessibility} % Improves PDF readability for those with disabilities.
\usepackage[pagenumbers]{cvpr} % To force page numbers, e.g. for an arXiv version
\usepackage{multirow} % Required for multirows

\definecolor{cvprblue}{rgb}{0.21,0.49,0.74}
\usepackage[pagebackref,breaklinks,colorlinks,allcolors=cvprblue]{hyperref}

\def\paperID{11352} % *** Enter the Paper ID here
\def\confName{CVPR}
\def\confYear{2025}

\title{Synergizing Motion and Appearance: Multi-Scale Compensatory Codebooks for Talking Head Video Generation}

\author{Shuling Zhao$^1$, \ \ Fa-Ting Hong$^1$, \ \ Xiaoshui Huang$^2$, \ \ Dan Xu$^{1}$\\
\vspace{-10pt}
\and
$^1$The Hong Kong University of Science and Technology, \quad $^2$Shanghai Jiao Tong University\\
{\tt\small \{szhaoax,fhongac\}@connect.ust.hk, huangxiaoshui@sjtu.edu.cn, danxu@cse.ust.hk}
}

\begin{document}
\maketitle

\begin{abstract}
Talking head video generation aims to generate a realistic talking head video that preserves the person’s identity from a source image and the motion from a driving video. Despite the promising progress made in the field, it remains a challenging and critical problem to generate videos with accurate poses and fine-grained facial details simultaneously. Essentially, facial motion is often highly complex to model precisely, and the one-shot source face image cannot provide sufficient appearance guidance during generation due to dynamic pose changes. To tackle the problem, we propose to jointly learn motion and appearance codebooks and perform multi-scale codebook compensation to effectively refine both the facial motion conditions and appearance features for talking face image decoding. Specifically, the designed multi-scale motion and appearance codebooks are learned simultaneously in a unified framework to store representative global facial motion flow and appearance patterns.~Then, we present a novel multi-scale motion and appearance compensation module, which utilizes a transformer-based codebook retrieval strategy to query complementary information from the two codebooks for joint motion and appearance compensation. The entire process produces motion flows of greater flexibility and appearance features with fewer distortions across different scales, resulting in a high-quality talking head video generation framework.
Extensive experiments on various benchmarks validate the effectiveness of our approach and demonstrate superior generation results from both qualitative and quantitative perspectives when compared to state-of-the-art competitors. \textcolor{black}{The project page is available at \href{https://shaelynz.github.io/synergize-motion-appearance/}{https://shaelynz.github.io/synergize-motion-appearance/}.}
%\footnote{\href{https://shaelynz.github.io/synergize-motion-appearance/}{https://shaelynz.github.io/synergize-motion-appearance/}}

\end{abstract}

%\footnote{\url{https://shaelynz.github.io/synergize-motion-appearance/}}    
\vspace{-3pt}
\section{Introduction}
\label{sec:intro}
\vspace{-2pt}
% \begin{figure}[tb]
% \centering
%   \includegraphics[width=1\linewidth]{figures/teaser.pdf}
%   \vspace{-10pt}
%   \caption{Motivation of our work.
%   }
%   \label{fig:teaser}
%   \vspace{-15pt}
% \end{figure}

\begin{figure}[tb]
\centering
  \includegraphics[width=1\linewidth]{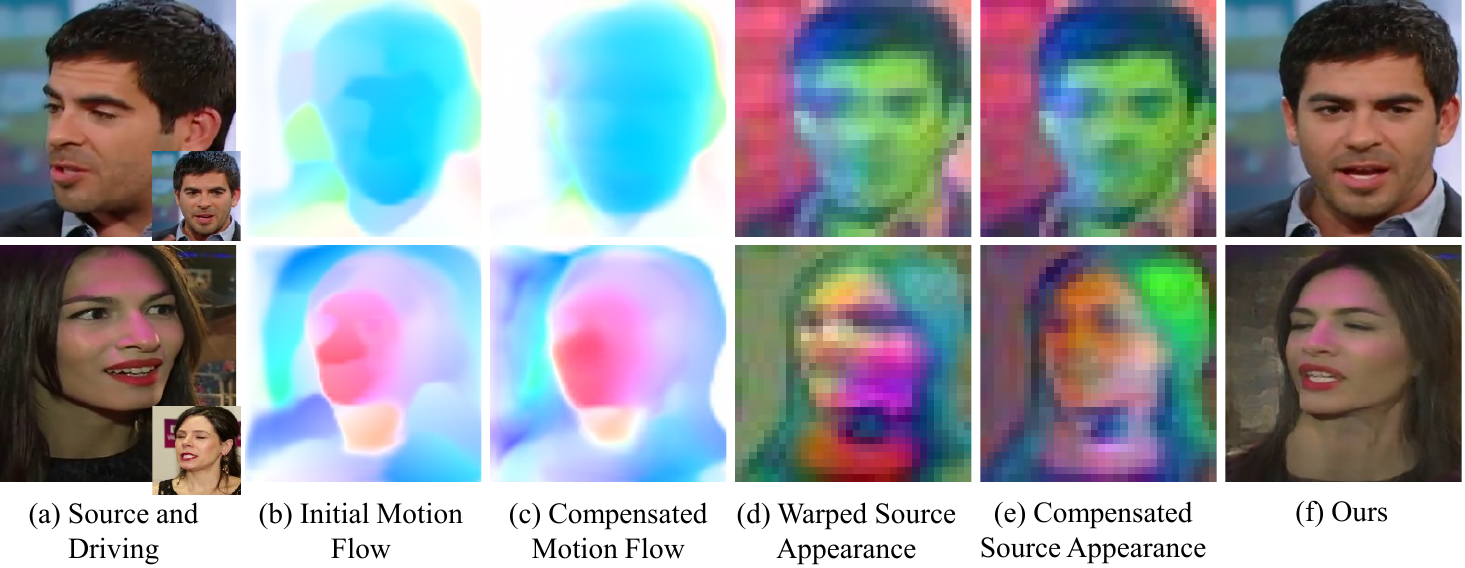} %{figures/teaser_compensate_update.pdf} %{figures/teaser_compensate_update_embed.pdf}
  \vspace{-15pt}
  \caption{Effect of motion and appearance compensation with jointly learned compensatory codebooks. Motion flows and warped source appearance features are refined by the complementary information retrieved from the motion and appearance codebooks, which jointly contribute to high-quality generated results.}
  \label{fig:teaser}
  \vspace{-15pt}
\end{figure}

Given a source image and a driving video, talking head video generation~\cite{hong2022depth, tao2024learning} aims to animate the person in the source image using the pose and expression from the driving video. Due to its widespread applications, such as video conferencing, the film industry, and virtual reality, it has attracted growing interest in the community.

%Significant progress has been made on this task in terms of both quality and robustness in recent years. Existing works primarily focus on learning more accurate motion estimation and appearance representation in 2D or 3D to enhance generation quality.
\textcolor{black}{Recent years have witnessed significant advancements in quality and robustness for this task. Current approaches primarily focus on improving motion estimation accuracy and appearance representation, whether in 2D or 3D, to enhance generation quality.}~Along the direction, unsupervised methods target predicting local motion flows around unsupervised keypoints without relying on facial priors~\citep{siarohin2019first, Zhao_2022_CVPR, wang2024continuous}, and methods based on predefined models (\emph{e.g.}, 3DMM)~\citep{Zakharov_2019_ICCV, Zhang_2023_CVPR, Ha_Kersner_Kim_Seo_Kim_2020} focus on learning robust decoding features to generate high-quality face outputs.~Despite the promising achievements, critical challenges persist: 1)~Some motion patterns (\emph{e.g.}, local subtle motions) cannot be inferred from a single image pair solely relying on unsupervised keypoints or predefined models for motion estimation, as such models often have limited power of motion representation and may fail to capture certain dynamic aspects of the facial motion from single image pairs (see Fig.~\ref{fig:teaser}b).
%, as static frames fail to capture certain dynamic aspects of the facial motion. Relying solely on unsupervised keypoints or predefined models leads to limitations in the accuracy of motion flow estimation, especially for complex, non-rigid motions. 
2) Even with accurate motion estimation, highly dynamic and complex motions in driving videos can create ambiguity during generation, as a still source image lacks sufficient appearance information to handle occluded regions or subtle expression changes (see Fig.~\ref{fig:teaser}d). This results in noticeable artifacts and a significant drop in the quality of the generated output. Therefore, generating realistic-looking facial images requires not only inferring accurate motion flow between the two given facial images but also compensating for the intermediate appearance decoding features from the one-shot source image for the final generation of face images.

In this work, we aim to synergize motion and appearance by simultaneously learning accurate motion flows for facial warping and robust facial appearance features for face image decoding, to advance talking head generation.
Inspired by the success of codebook learning~\citep{van2017neural} where compact and useful representations of certain modalities are learned from the whole training dataset, we propose to use such representations as additional knowledge and compensate for motion and appearance with them. 
Specifically, we propose a unified framework that can achieve joint learning of both motion and appearance codebooks with multi-scale compensation. 
To refine the motion flow between two facial images (\emph{i.e.}, source and driving), we design a multi-scale motion codebook that captures diverse motion patterns across scales from the entire dataset during training. 
%Using this learned multi-scale motion codebook, we further devise a transformer-based compensation structure to iteratively refine motion flows in a coarse to fine manner. 
To enhance intermediate warped facial feature maps for image decoding, we introduce a multi-scale appearance codebook that represents diverse appearance patterns learned from the entire dataset. 
To use the motion and appearance codebooks, we further introduce a transformer-based compensation structure to iteratively refine motion flows in a coarse to fine manner and refine the warped features across different scales. This approach enables us to improve motion accuracy and capture more facial details by leveraging the diverse motion and appearance information stored in the codebooks.
To enhance the learning and compensation of both codebooks, we propose a joint training strategy in which the motion and appearance codebooks are learned simultaneously with the entire framework. This approach allows both codebooks to be optimized together, utilizing gradients from the refined warped features to strengthen their mutual influence and improve overall performance.
By learning both multi-scale motion and appearance codebooks, our framework refines the motion flow to accurately warp the source facial features, which are further compensated with additional details from the appearance codebook. This process yields robust intermediate facial decoding features, resulting in improved generation.

We conduct extensive ablation studies to verify the effectiveness of the learned multi-scale motion and appearance codebooks. Experimental results demonstrate that both codebooks effectively enhance the motion flow and intermediate warped features, resulting in more accurate and detailed facial motion flows and feature textures.~Furthermore, results on two challenging datasets indicate that our method surpasses state-of-the-art approaches, producing realistic-looking talking head videos. In summary, our contributions are threefold:
 % \vspace{-5pt}
\begin{itemize}[leftmargin=*]
    \item We propose a novel framework that \emph{jointly learns multi-scale motion and appearance codebooks}. The motion codebook captures motion patterns at varying levels of granularity, while the appearance codebook stores representative facial structure and texture features. This joint learning enables the model to effectively compensate \textcolor{black}{for} both motion and appearance for advanced generation.
    
    \item We develop an effective \emph{multi-scale compensation mechanism} that utilizes the learned motion and appearance codebooks to progressively refine both motion and appearance representations. The mechanism can couple the compensation of both aspects at each level, achieving higher consistency of appearance and motion, thus leading to high visual quality in generated videos. 
    % ensures more accurate motion estimation and thus enhances the visual quality of generated videos.     
    % the precision of motion estimation, particularly in complex scenarios, 
    % and enhances the visual quality by enriching the texture details in the generated video.
    \item Extensive experiments\textcolor{black}{,} including \textcolor{black}{those} on challenging datasets\textcolor{black}{,} demonstrate that our method not only effectively compensates for facial motions and appearances but also significantly outperforms state-of-the-art approaches, generating more realistic talking head videos.
\end{itemize}

\begin{figure*}[tb]
    \centering
  \includegraphics[width=1\linewidth]{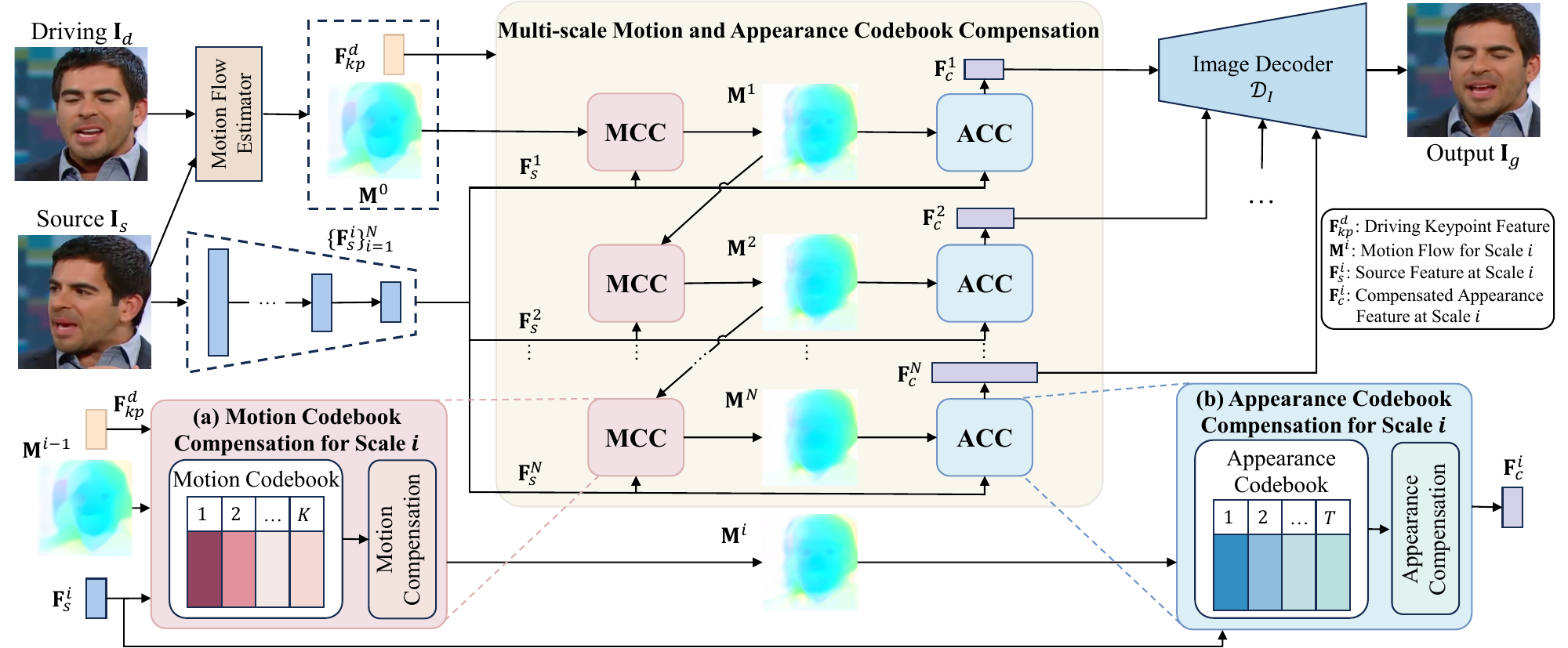}
  \vspace{-10pt}
  \caption{Overview of the framework. 
  For each scale, multi-scale motion and appearance codebook compensation consists of two sub-modules. (i) \textbf{Motion Codebook Compensation (MCC)} compensates for a motion flow with the motion codebook. (ii) To refine the source facial feature warped by the compensated motion flow, \textbf{Appearance Codebook Compensation (ACC)} produces the compensated appearance feature with the appearance codebook for image decoding. These two sub-modules are employed for all scales. We learn the motion and appearance codebooks jointly with the whole framework. 
  }
  \label{fig:overview}
  \vspace{-15pt}
\end{figure*}

\vspace{-2pt}
\section{Related Work}
\label{sec:related_work}
\vspace{-3pt}
\noindent\textbf{Talking Head Video Generation.} 
Existing works on talking head video generation generally separate the motion estimation module and \textcolor{black}{the} image generation module to disentangle appearance and motion. To transfer the motion, some works require facial priors provided by a pre-trained model during generation. For example, landmark-based approaches \citep{Ha_Kersner_Kim_Seo_Kim_2020, Zhao_2021_ICCV, Wu_2018_ECCV, 10.1007/978-3-030-58610-2_31} detect \textcolor{black}{predefined} facial landmarks to transfer the facial pose and expression from a driving frame to the source image. Some other methods \citep{Ren_2021_ICCV, Zeng_2023_CVPR, yao2020mesh} use parameters from 3D face models \citep{blanz2023morphable, feng2021learning, zhu2017face} as motion descriptors to disentangle identity and pose. However, they normally cannot describe non-facial parts such as hair and neck, and their generation quality is limited by the pre-trained model performance. To address the issue, several methods that do not require any prior knowledge from pre-trained models are proposed. Monkey-Net \citep{siarohin2019animating} learns sparse motion-related keypoints in an unsupervised manner to describe object movements. FOMM \citep{siarohin2019first} extends it with local affine transformation assumption around the keypoints to model complex motion. Subsequent works introduce more flexible mathematical models such as thin-plane spline transformation \citep{Zhao_2022_CVPR} and continuous piecewise-affine-based transformation \citep{wang2024continuous} to increase motion estimation accuracy. Despite the expressiveness of the motion models, they cannot fully describe large head poses and delicate expression changes. MRFA \citep{tao2024learning} tackles the problem by building a correlation volume for each image pair and using it to refine the coarse motion flow iteratively. However, it only uses the warped image feature and a plain image generator for image generation, which may fail when facing extreme pose change\textcolor{black}{s,} as the appearance information from the one-shot source image is \textcolor{black}{often insufficient}. Some works \citep{yin2022styleheat, Oorloff_2023_ICCV, Bounareli_2023_ICCV} leverage the remarkable generative power of pre-trained StyleGAN \citep{Karras_2019_CVPR, Karras_2020_CVPR} for better image generation, but they usually have to balance the editability and fidelity. MCNet \citep{Hong_2023_ICCV} learns a meta-memory bank of spatial facial features to compensate for the warped source features. Different from previous works, we compensate for both motion flows and warped source features with jointly learned multi-scale motion and appearance codebooks to boost the generation quality.

\noindent\textbf{Codebook Learning.} 
Codebook learning aims to learn useful discrete representations with a fixed size. The learned codebook contains rich and compact information, which can facilitate various tasks such as image classification \citep{cai2010learning, zhang2009learning}, image synthesis \citep{esser2021taming, Chang_2022_CVPR}, blind face restoration \citep{zhou2022towards, gu2022vqfr} and audio-driven talking head video generation \citep{Wang_2023_CVPR_LipFormer}. Traditional methods often learn a codebook with unsupervised clustering such as k-means \citep{csurka2004visual}. % and random forests \citep{moosmann2006fast}. 
VQ-VAE \citep{van2017neural} first incorporates vector quantization in a Variational Autoencoder to learn a codebook containing discrete representative input features. To build a context-rich codebook for images, VQGAN \citep{esser2021taming} further increases the compression rate and adds a discriminator and a perceptual loss on images reconstructed with the codes from the codebook. A transformer is later used to model the composition of the codes for high-resolution image synthesis. CodeFormer \citep{zhou2022towards} uses a learned codebook of compressed high-quality face image features as discrete prior and predicts the code sequence based on the low-quality facial input for blind face restoration. LipFormer \citep{Wang_2023_CVPR_LipFormer} learns two codebooks of the upper half face and the bottom half face respectively and predicts the lip codes from the input audio to generate a face video from the audio. We also adopt the idea of codebook learning, but we simultaneously learn multi-scale motion and appearance codebooks that store diverse motion and appearance patterns from the entire dataset during training to facilitate high-quality talking head video generation. The codebooks and the entire framework are trained together so that patterns useful for talking head video generation can be stored in and retrieved from the codebooks. 
% \vspace{-10pt}
\section{The Proposed Method}
% \vspace{-5pt}
% \begin{figure*}[tb]
%     \centering
%   \includegraphics[width=1\linewidth]{figures/overview_2_4_1.pdf}
%   \vspace{-10pt}
%   \caption{Overview of the framework.
%   For each scale, it consists of two sub-modules. (i) \textbf{Motion Codebook Compensation (MCC)} compensates for a motion flow with the motion codebook. (ii) To compensate for the source facial feature warped by the compensated motion flow, \textbf{Appearance Codebook Compensation (ACC)} uses the appearance codebook and produces the compensated appearance feature for generation. These two sub-modules are employed for all scales. We learn the motion and appearance codebook jointly with the whole framework. 
%   }
%   \label{fig:overview}
% \end{figure*}

% \begin{figure}[t]
%   \begin{center}
%   \includegraphics[width=1\linewidth]{figures/motion_compensation_update.pdf}
%   \end{center}
%   \vspace{-15pt}
%   \caption{Illustration of motion codebook learning and compensation for scale $i$. We adopt a transformer structure $\mathcal{T}_M$ to compensate for the motion flow using the learned motion codebook.~The proposed motion codebook is learned under the supervision of a reconstruction loss and a code-level loss.}
%   \label{fig:motion_comp}
%   \vspace{-20pt}
  
% \end{figure}

In this section, we will present the details of our framework. We jointly learn multi-scale motion and appearance codebooks with compensation for the motion and intermediate appearance features during generation. We adopt the Taylor expansion approximation method to learn the initial motion flow and the warping manner as same as \citet{siarohin2019animating} for video generation.
% The proposed multi-scale motion and appearance codebook compensation-based framework is illustrated in Fig. \ref{fig:overview}. We adopt the frame-to-frame video generation pipeline in \citep{siarohin2019animating} and take the source image $\mathbf{I}_s$ and the $t^{th}$ driving frame $\mathbf{I}_d$ as input at time $t$. \fating{**TODO: claim your main idea here**}

% \vspace{-5pt}
\subsection{Overview}

Our framework is illustrated in Fig.~\ref{fig:overview}. First, a keypoint-based motion flow estimator takes both the source image $\mathbf{I}_s$ and driving image $\mathbf{I}_d$ as input and estimates the initial coarse motion flow $\mathbf{M}^0$. An image encoder $\mathcal{E}_I$ extracts multi-scale source features $\{\mathbf{F}_s^i\}_{i=1}^N$ from $\mathbf{I}_s$. Using $\mathbf{M}^0$, $\{\mathbf{F}_s^i\}_{i=1}^N$, and the driving keypoint feature $\mathbf{F}_{kp}^d$, the multi-scale motion and appearance codebook compensation module refines the motion flows and warped source features across all scales to obtain the multi-scale compensated appearance features $\{\mathbf{F}_c^i\}_{i=1}^N$ with more accurate motion and less distortion. 
Finally, the image decoder $\mathcal{D}_I$ decodes $\{\mathbf{F}_c^i\}_{i=1}^N$ to generate the final image $\mathbf{I}_g$ with the target motion and appearance. 
%Therefore, through multi-scale motion and appearance codebook compensation, 
Details on the design and learning of multi-scale motion and appearance codebooks and compensation are provided in the following subsections.

\subsection{Multi-scale Motion and Appearance Codebook Learning and Compensation}
\label{subsec:joint_cdbk}
% \vspace{-5pt}

\begin{figure}[tb]
  \begin{center}
  \includegraphics[width=1\linewidth]{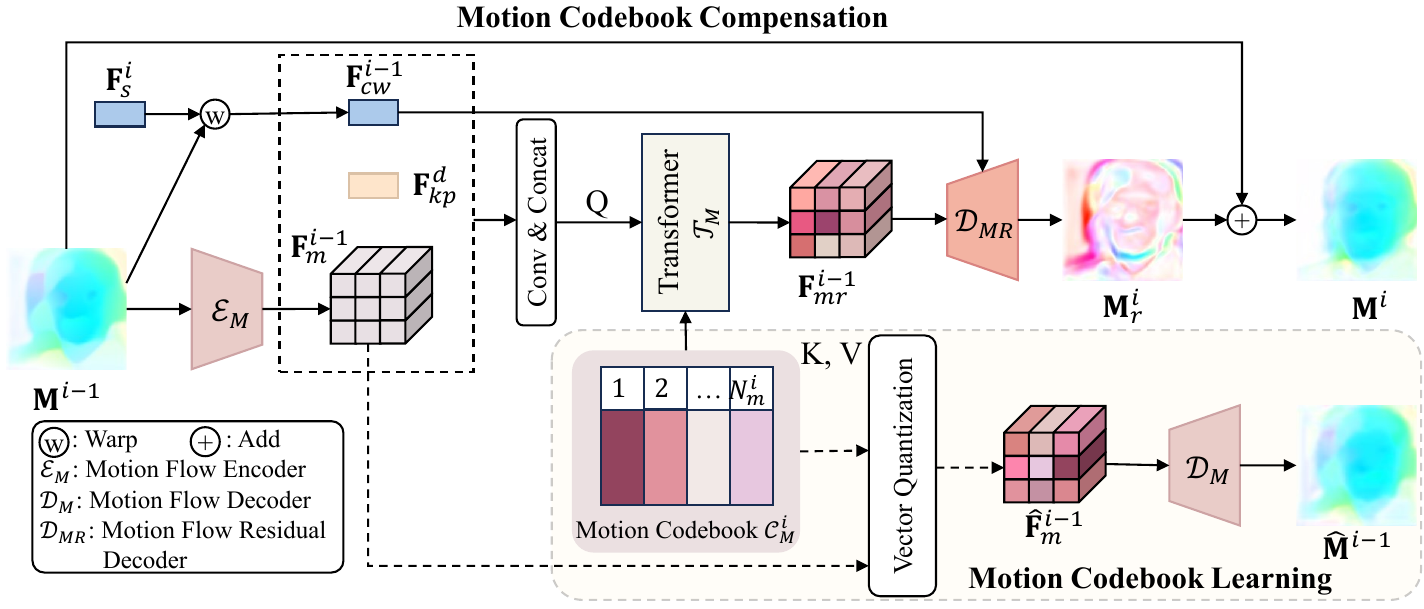}
  \end{center}
  \vspace{-20pt}
  \caption{Illustration of motion codebook learning and compensation for scale $i$. We adopt a transformer structure $\mathcal{T}_M$ to compensate for the motion flow using the motion codebook.~The proposed motion codebook is learned under the supervision of a reconstruction loss and a code-level loss.}
  \label{fig:motion_comp}
  \vspace{-15pt}
  
\end{figure}

\begin{figure}[tb]
  \begin{center}
  \includegraphics[width=1\linewidth]{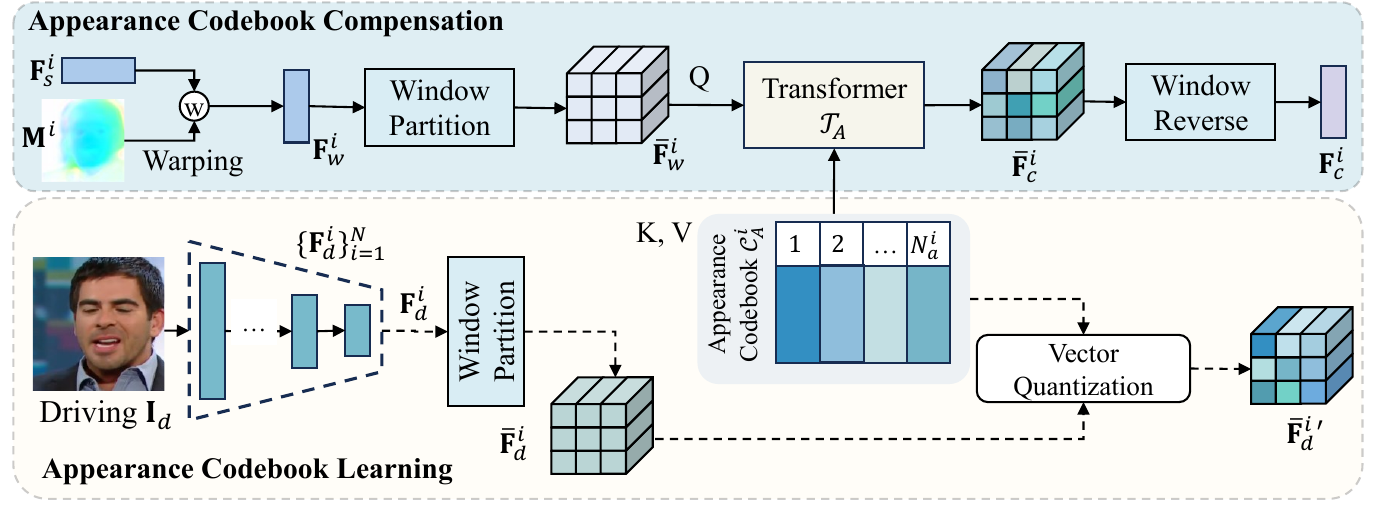}
  \end{center}
  \vspace{-20pt}
  \caption{Illustration of appearance codebook learning and compensation for scale $i$. We utilize a transformer structure $\mathcal{T}_A$ to compensate for the warped features with the appearance codebook, adding more facial details to the feature map. Similar to the motion codebook, the designed appearance codebook is learned under the supervision of a code-level loss.}
  \label{fig:app_comp}
  \vspace{-15pt}
\end{figure}

%Previous methods typically estimate motion flows with source and driving features from a fixed scale with certain mathematical models. While they capture rough motion, their accuracy is limited by the single-scale information and the assumed transformations around unsupervised keypoints, especially in complex motion scenarios. Meanwhile, even with accurate motion, when the motion between source and driving images is too large, occlusion can cause the warped features to lack sufficient appearance details. 
We refine the initial motion flow $\mathbf{M}^0$ from coarse to fine and enhance multi-scale source features warped by the refined motion flows with the multi-scale motion and appearance codebook compensation module, producing more accurate motion flows for source feature warping at different scales and restore natural human faces from warping distortions, which together result in higher-quality facial decoding features. Specifically, we jointly learn a multi-scale motion codebook storing local motion flow patterns and a multi-scale appearance codebook containing local facial textures and retrieve relevant information from them. Fig.~\ref{fig:motion_comp} and Fig.~\ref{fig:app_comp} illustrate the motion and appearance codebook learning and compensation process for scale $i$ respectively.

%using a multi-scale motion codebook and enhance multi-scale source features warped by the refined motion flows with a multi-scale appearance codebook. This enables us to produce more accurate motion flows for source feature warping at different scales and restore natural human faces from warping distortions. Specifically, we jointly learn a multi-scale motion codebook storing local motion flow patterns and a multi-scale appearance codebook containing local facial textures and retrieve relevant information from them. Fig.~\ref{fig:motion_comp} and Fig.~\ref{fig:app_comp} illustrate the motion and appearance codebook learning and compensation process for scale $i$ respectively.

\subsubsection{Multi-scale Code Allocation}
\begin{figure}[tb]
  \begin{center}
  %\vspace{-13pt}
  \includegraphics[width=1\linewidth]{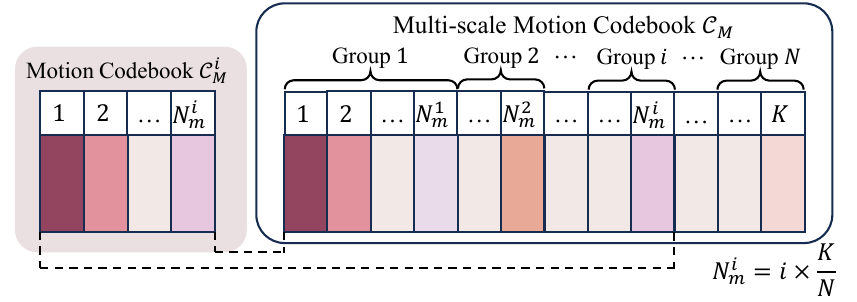}
  \end{center}
  \vspace{-20pt}
  \caption{Illustration of the code allocation scheme. We take the multi-scale motion codebook as an example.}
  \label{fig:code_allocation}
  \vspace{-15pt}
\end{figure}

% \vspace{-5pt}
We aim to refine the initial motion flow and the warped source features across all $N$ scales with multi-scale motion and appearance codebooks. As the scale increases, larger appearance features require finer motion flows for accurate warping and more detailed appearance information for appearance compensation. To provide sufficient motion and appearance compensation at larger scales, we introduce a code allocation scheme for multi-scale motion and appearance codebooks, which divides the codebook into multiple groups and allocates more codes for larger scales. We illustrate the scheme with the motion codebook in Fig.~\ref{fig:code_allocation}. The motion codebook $\mathcal{C}_M=\{\mathbf{m}_k\in \mathbb{R}^{d_m}\}_{k=1}^{K}$ contains $K$ codes, and we perform motion compensation at $N$ different scales. The $K$ codes are split into $N$ equal groups, each with $K/N$ codes. At scale $i$, the first $i$ groups, totaling $N_m^i = i \times K/N$ codes, are allocated for motion compensation. Similarly, for the appearance codebook $\mathcal{C}_A=\{\mathbf{a}_k\in \mathbb{R}^{d_a}\}_{k=1}^{T}$ containing $T$ codes, we also allocate the first $i$ groups, which are the first $N_a^i = i \times T/N$ codes for appearance compensation.
This allows codes with smaller indices to capture general motion patterns or coarse appearance patterns shared across scales, while codes with larger indices to focus on finer motion patterns or facial details needed for larger scales. This maximizes the use of the two codebooks by sharing general information across scales while reserving space for scale-specific details. 
%Thus, codes with smaller indices capture general, coarse appearance patterns shared across scales, while larger-index codes focus on finer facial details. 

At each scale $i$, we form new motion and appearance codebooks $\mathcal{C}_M^i$ and $\mathcal{C}_A^i$ from the allocated codes, resulting in $N$ scale-specific motion codebooks $\{\mathcal{C}_M^i\}_{i=1}^N$ and $N$ scale-specific appearance codebooks $\{\mathcal{C}_A^i\}_{i=1}^N$ for multi-scale motion and appearance compensation.

% \vspace{-5pt}
\subsubsection{Joint Codebook Learning}
To realize effective motion and appearance compensation at different scales, we directly update $\{\mathcal{C}_M^i\}_{i=1}^N$ and $\{\mathcal{C}_A^i\}_{i=1}^N$ with motion flow and appearance feature units at each scale. We jointly optimize the two codebooks with the whole framework, allowing the network to effectively store and retrieve useful patterns at different scales with the codebooks, leading to high-quality talking head video generation.
%so that only motion and appearance patterns useful for talking head video generation are stored.

% for effective compensation at different scales, we directly update $\{\mathcal{C}_M^i\}_{i=1}^N$ and $\{\mathcal{C}_A^i\}_{i=1}^N$.
%To learn motion and appearance codebooks storing multi-scale motion flow patterns and multi-scale appearance codes for effective compensation at different scales, we directly update $\{\mathcal{C}_M^i\}_{i=1}^N$ and $\{\mathcal{C}_A^i\}_{i=1}^N$.
%following the idea of vector quantization \citep{van2017neural}.

\noindent\textbf{Motion Codebook Learning.}
At scale $i$, we use a CNN-based motion flow encoder $\mathcal{E}_M$ to map the input motion flow $\mathbf{M}^{i-1} \in \mathbb{R}^{h \times w \times 2}$ into a compact motion flow feature $\mathbf{F}_{m}^{i-1}\in \mathbb{R}^{h_m \times w_m \times d_m}$ where each unit is of length $d_m$ and captures a local motion flow pattern on $\mathbf{M}^{i-1}$. %To update the motion codebook with such local motion flow patterns, 
We quantize each of its spatial element\textcolor{black}{s} $\mathbf{F}_{m}^{i-1}(x,y)$ with \textcolor{black}{the} nearest code from $\mathcal{C}_M^i$ to obtain a quantized motion flow feature $\hat{\mathbf{F}}_{m}^{i-1}$:
%As shown in Fig. \ref{fig:motion_comp} (a), following idea of vector quantization, we adopt a motion flow encoder $E_M$, a motion flow decoder $\mathcal{D}_M$ and a motion codebook $_M=\{c_M^k\in \mathbb{R}^{d_m}\}_{k=1}^{d_m}$. We encode the coarse motion flow $M_c \in \mathbb{R}^{h \times w \times 2}$ with a CNN-based motion flow encoder $E_M$ into a compact coarse motion flow feature $F_{Mc}\in \mathbb{R}^{h_m \times w_m \times d_m}$, each unit of which of length $d_m$ captures a local motion flow pattern. To update the motion codebook with such local motion flow patterns, we quantize each element of $F_{Mc}$ with the nearest codes from the motion codebook to produce a quantized motion feature $\hat{F}_{Mc}$:
\begin{equation}
\hat{\mathbf{F}}_{m}^{i-1} = \mathit{Q}(\mathbf{F}_{m}^{i-1}) := \left( \mathop{\arg\min}_{\mathbf{m}_k\in \mathcal{C}_M^i} ||\mathbf{F}_{m}^{i-1}(x,y)-\mathbf{m}_k||_2^2 \right). %\in \mathbb{R}^{h_m \times w_m \times d_m}.
\end{equation}
A motion flow decoder $\mathcal{D}_M$ reconstructs $\mathbf{M}^{i-1}$ with the quantized motion flow feature $\hat{\mathbf{F}}_{m}^{i-1}$:
\begin{equation}
\hat{\mathbf{M}}^{i-1} = \mathcal{D}_M(\hat{\mathbf{F}}_{m}^{i-1})= \mathcal{D}_M(\mathit{Q}(\mathcal{E}_M(\mathbf{M}^{i-1}))).
\end{equation}
To update the scale-specific motion codebook $\mathcal{C}_M^i$ with local motion flow patterns from $\mathbf{F}_{m}^{i-1}$, we use the following loss function:
\begin{equation}
\begin{split}
\label{l_vqm}
%\hat{M}_{c} = D_M(\hat{F}_{Mc})= D_M(Q(E_m(M_c)))
\mathcal{L}_{vq,m}^i = &\lambda _{recon,m}||\hat{\mathbf{M}}^{i-1}-sg[\mathbf{M}^{i-1}]||_1 \\
&+ ||sg[\mathcal{E}_M({\mathbf{M}}^{i-1})]-\hat{\mathbf{F}}_{m}^{i-1}||_2^2 \\ 
& + \beta ||sg[\hat{\mathbf{F}}_{m}^{i-1}]-\mathcal{E}_M(sg[\mathbf{M}^{i-1}])||_2^2,
\end{split}
\end{equation}
% where $sg[\cdot]$ denotes the stop gradient operator, and $\lambda _{recon,m}$ and $\beta$ denote the weights of the loss terms. The first term is the motion flow reconstruction loss, while the last two terms are the losses to close the distance between the latent motion flow units and the motion codes. We stop the gradient of motion codebook training for $\mathbf{M}^{i-1}$ so that the training of the motion flow estimator or the compensated motion flow from the prior scales is not affected by the motion codebook learning of scale $i$. Considering all $N$ scales, the overall loss function for multi-scale motion codebook learning is $\mathcal{L}_{vq,m} = \sum _{i=1}^N \mathcal{L}_{vq,m}^i$.
where $sg[\cdot]$ denotes the stop gradient operator, and $\lambda_{recon,m}$ and $\beta$ are the loss term weights. The first term represents the motion flow reconstruction loss, while the last two terms form a code-level loss \citep{van2017neural} that minimizes the distance between the latent motion flow units and the motion codes. We stop the gradient of $\mathbf{M}^{i-1}$ to ensure that motion codebook learning at scale $i$ does not interfere with the training of the motion flow estimator or the compensated motion flows from prior scales. The overall loss function for motion codebook learning across all $N$ scales is $\mathcal{L}_{vq,m} = \sum_{i=1}^N \mathcal{L}_{vq,m}^i$.

\noindent\textbf{Appearance Codebook Learning.}
We use the image encoder $\mathcal{E}_I$ to extract multi-scale driving features $\{\mathbf{F}_d^i\}_{i=1}^N$, which serve as targets for our appearance codebook. Since image features of different scales have varying resolutions, directly flattening high-resolution features for compensation can be computationally expensive. To address this, we apply window partitioning. For a feature map of shape $(h_a^i, w_a^i, c_a^i)$ at scale $i$, we divide it into patches of shape $(h_a^i/h_a, w_a^i/w_a)$, reshaping the feature into $(h_a, w_a, c_a^i \times h_a^i \times w_a^i / h_a / w_a)$. We then linearly project the features into $d_a$ dimensions to align them with the appearance codes. This results in a more compact driving feature $\overline{\mathbf{F}}_d^i\in \mathbb{R}^{h_a\times w_a\times d_a}$, where each unit represents an appearance pattern at scale $i$. Finally, element-wise quantization with the nearest code from $\mathcal{C}_A^i$ produces the quantized appearance feature $\overline{\mathbf{F}}_d^{i\prime}$:
\begin{equation}
\overline{\mathbf{F}}_d^{i \prime} = \mathit{Q}(\overline{\mathbf{F}}_d^i) := \left( \mathop{\arg\min}_{\mathbf{a}_k\in \mathcal{C}_A^i} ||\overline{\mathbf{F}}_d^i(x,y)-\mathbf{a}_k||_2^2 \right). % \in \mathbb{R}^{h_a \times w_a \times d_a}.
\end{equation}
To update the scale-specific appearance codebook $\mathcal{C}_A^i$ with local appearance units from $\overline{\mathbf{F}}_d^i$, we use the following loss function:
\begin{equation}
\label{l_vqa}
\mathcal{L}_{vq,a}^i = ||sg[\overline{\mathbf{F}}_d^i]-\overline{\mathbf{F}}_d^{i\prime}||_2^2  + \beta ||sg[\overline{\mathbf{F}}_d^{i\prime}]-\overline{\mathbf{F}}_d^i||_2^2,
\end{equation}
where $sg[\cdot]$ denotes the stop gradient operator, and $\beta$ is the loss term weight. We only use the code-level loss to reduce the distance between the appearance feature units and the codes from $\mathcal{C}_A^i$. The overall loss function for appearance codebook learning is $\mathcal{L}_{vq,a} = \sum_{i=1}^N \mathcal{L}_{vq,a}^i$. We do not use the image decoder $\mathcal{D}_I$ to reconstruct the original driving image $\mathbf{I}_d$ from $\{\overline{\mathbf{F}}_d^{i\prime}\}_{i=1}^N$, allowing $\mathcal{D}_I$ to focus on decoding the compensated appearance features and improving the talking head video generation. Training a separate image decoder for reconstruction with quantized appearance codes would be computationally expensive. Experimental results in Sec.~\ref{subsec:exp-ablation} show that using only the code-level loss can already learn the appearance codebook effectively.

\subsubsection{Multi-scale Joint Codebook Compensation} 
\par\noindent\textbf{Motion and Appearance Codebook Compensation Coupling.}
With the multi-scale motion and appearance codebooks, we couple motion and appearance codebook compensation at each scale to produce better image decoding features, as shown in Fig.~\ref{fig:overview}. Specifically, at \textcolor{black}{scale $i$}, motion codebook compensation (MCC) refines the input motion flow $\mathbf{M}^{i-1}$ to obtain the compensated motion flow $\mathbf{M}^i$, which is used to warp the source feature $\mathbf{F}_s^i$. Appearance codebook compensation (ACC) then refines the warped source feature to produce the compensated appearance feature $\mathbf{F}_c^i$. If $i<N$, $\mathbf{M}^{i}$ is used as the input motion flow for the next scale. In this way, motion and appearance codebook compensation are coupled across scales, enabling more consistent motion and appearance for high-quality video generation.

\noindent\textbf{Motion Codebook Compensation.}
An intuitive approach for multi-scale motion codebook compensation is to retrieve motion codes from the scale-specific codebook and decode them into finer motion flows using $\mathcal{D}_M$ at each scale. However, to reduce computational complexity, we limit the number of motion codes, which decreases expressiveness. This makes it difficult to fully reconstruct motion flows, leading to degraded image quality. Instead, we predict motion flow residual for the input flow at each scale, allowing the motion codebook to enhance the motion flow without requiring precise reconstruction.

As shown in Fig.~\ref{fig:motion_comp}, to retrieve motion residuals from $\mathcal{C}_M^i$ at scale $i$, we use a motion code retrieval transformer $\mathcal{T}_M$ shared for all scales. It queries the scale-specific codebook $\mathcal{C}_M^i$ using the encoded motion feature $\mathbf{F}_m^{i-1}$, the warped source feature $\mathbf{F}_{cw}^{i-1}$, and the driving keypoint feature $\mathbf{F}_{kp}^d$. The first two represent the current motion flow, and the last indicates the target pose. These features are processed through a convolutional encoding block and concatenated into a compact motion query feature, then flattened and enhanced with a learnable position embedding before being passed to $\mathcal{T}_M$. $\mathcal{T}_M$ consists of $L_M$ transformer layers, each with multi-head self-attention, cross-attention, and convolution layers (instead of linear layers) to preserve spatial structure. The self-attention models global correlations, and the cross-attention uses the output of self-attention as the query and the codes from $\mathcal{C}_M^i$ as key-value pairs to retrieve local motion flow patterns. The transformer outputs the motion flow residual feature $\mathbf{F}_{mr}^{i-1}$, which is decoded by a motion flow residual decoder $\mathcal{D}_{MR}$ to obtain the motion flow residual $\mathbf{M}_r^i \in \mathbb{R}^{h \times w \times 2}$. Finally, we add $\mathbf{M}_r^i$ to the input motion flow $\mathbf{M}^{i-1}$ to produce the compensated motion flow $\mathbf{M}^i$, which is used for source feature warping at scale $i$.

The compensated motion flow $\mathbf{M}^i$ is sufficient for warping the source feature $\mathbf{F}_s^i$, but may lack fine details for higher-resolution features like $\mathbf{F}_s^{i+1}$. Therefore, we use $\mathbf{M}^i$ as input for motion codebook compensation at scale $i+1$ and refine it with more motion codes. This iterative process continues across scales, producing multi-scale compensated motion flows $\{\mathbf{M}^i\}_{i=1}^N$.

%%%%%% appearance
\noindent\textbf{Appearance Codebook Compensation.}
%To transfer the compensated motion to the source image, we warp the source features with their corresponding compensated motion flows at each scale. However, warping can introduce distortions due to pose changes, which degrade image quality. To address this, we use the multi-scale appearance codebook to repair the distorted warped features with retrieved appearance information. 
At scale $i$, we first warp the source feature $\mathbf{F}_s^i$ with the compensated motion flow $\mathbf{M}^i$ to obtain the warped feature $\mathbf{F}_w^i$. We then apply window partitioning to map $\mathbf{F}_w^i$ into a compact feature $\overline{\mathbf{F}}_w^i$. To correct the corrupted appearance in $\overline{\mathbf{F}}_w^i$, we retrieve appearance codes with $\overline{\mathbf{F}}_w^i$ using a transformer $\mathcal{T}_A$.
Similar to motion codebook compensation, $\overline{\mathbf{F}}_w^i$, reshaped and augmented with a learnable position embedding, passes through $L_A$ transformer layers where appearance codes from $\mathcal{C}_A^i$ are retrieved with cross-attention.
%, which also use convolution layers instead of linear layers to preserve spatial structures. The self-attention mechanism models global interactions, and cross-attention retrieves appearance codes from $\mathcal{C}_A^i$. 
The transformer output is the compensated appearance feature $\overline{\mathbf{F}}_c^i$, which retains the pose but reduces distortion of $\overline{\mathbf{F}}_w^i$. Finally, we apply window reverse on $\overline{\mathbf{F}}_c^i$ to restore it to the original image shape $(h_a^i, w_a^i, c_a^i)$ using linear projection and reshaping. This compensation is performed across all scales, resulting in multi-scale compensated appearance features $\{\mathbf{F}_c^i\}_{i=1}^N$.

%%%%%%%%%%%%%%%%%% Reference Line

%%%%%%%%%%%%%EEEEEENDDDDDDDDDD

% \vspace{-5pt}
% \subsection{Motion Codebook Learning and Compensation}
% \label{subsec:motion_cdbk}
% \vspace{-5pt}

% \noindent\textbf{Multi-scale Motion Codebook Compensation.} 

% \vspace{-5pt}
% \subsection{Appearance Codebook Learning and Compensation}
% \label{subsec:app_cdbk}
% \vspace{-5pt}

% We have obtained compensated motion flows for each scale and can warp the corresponding source features with better accuracy. However, when the motion between the source and driving images is too large, the warped source features cannot offer sufficient appearance information due to occlusion. To alleviate the problem, we compensate for the warped source features at each scale with appearance codebook compensation. Specifically, we learn a multi-scale appearance codebook consisting of local image textures and retrieve proper appearance information from it to enhance the warped source features for image generation. Fig. \ref{fig:app_comp} depicts the appearance codebook compensation for scale $i$.

% \vspace{-15pt}
\subsection{Joint Optimization Objective of the Framework}
\label{subsec:train}
% \vspace{-5pt}

We use the multi-scale compensated appearance features $\{\mathbf{F}_c^i\}_{i=1}^N$ and the image decoder $\mathcal{D}_I$ for image decoding. The low-resolution feature $\mathbf{F}_c^1$ is fed as the initial input to $\mathcal{D}_I$, which gradually upsamples it through a series of upsampling layers and ResNet blocks \citep{He_2016_CVPR} until it reaches the output resolution. The intermediate features are refined with higher-resolution features $\{\mathbf{F}_c^i\}_{i=2}^N$ by SFT \citep{wang2018recovering} and addition when they match the resolution of $\mathbf{F}_c^i$.
%, we apply SFT \citep{wang2018recovering} to refine the intermediate features when they match the resolution of $\mathbf{F}_c^i$. 
%We also pass $\mathbf{F}_c^i$ through a convolution layer and add to the intermediate features. 
After fusing \textcolor{black}{features from all scales}, $\mathcal{D}_I$ generates the final output $\mathbf{I}_g$.

The training objective combines the codebook losses from Sec.~\ref{subsec:joint_cdbk} with common losses for talking head video generation \citep{siarohin2019first, hong2022depth}. Specifically, we use the equivariance loss $\mathcal{L}_{eq}$ and keypoint distance loss $\mathcal{L}_{kpd}$ to guide keypoint prediction in the motion flow estimator, and an image reconstruction loss $\mathcal{L}_{recon}$
%L1 loss $\mathcal{L}_1$, perceptual loss $\mathcal{L}_p$, 
and an adversarial loss $\mathcal{L}_{adv}$ on $\mathbf{I}_g$. To avoid the image decoder relying too much on high-resolution appearance features, we also generate an image $\mathbf{I}_g^1$ using only $\mathbf{F}_c^1$ and apply $\mathcal{L}_{recon}$
%$\mathcal{L}_1$ and $\mathcal{L}_p$ 
to minimize the difference between $\mathbf{I}_g^1$ and $\mathbf{I}_d$. The overall training objective is:
\begin{align*}
\begin{split}
\mathcal{L} = &  \mathcal{L}_{recon}(\mathbf{I}_d, \mathbf{I}_g) + \lambda_{adv} \mathcal{L}_{adv}(\mathbf{I}_d, \mathbf{I}_g)  +  \mathcal{L}_{eq} + \mathcal{L}_{kpd} \\
& +  \mathcal{L}_{vq,m} + \mathcal{L}_{vq,a} + \lambda^1 \mathcal{L}_{recon}(\mathbf{I}_d, \mathbf{I}_g^1) ,
\end{split}
\end{align*}
where $\lambda_{adv}$ and $\lambda^1$ are \textcolor{black}{the loss} term weights.

%\vspace{-15pt}
\section{Experiments}
\label{sec:exp}
\vspace{-2pt}
\subsection{Implementation Details}
% \vspace{-5pt}
\noindent\textbf{Datasets.}
We conduct experiments on VoxCeleb1 \citep{nagrani2017voxceleb} and CelebV-HQ \citep{zhu2022celebv} datasets. We train our model on VoxCeleb1 training set. For evaluation, we build the test set on VoxCeleb1 by randomly sample 50 videos from its test split. To evaluate the model's generalization ability, we randomly select 50 videos from CelebV-HQ for testing. 
%For evaluation on CelebV-HQ, we randomly select 200 videos and sample 2000 image pairs from them.

\noindent\textbf{Metrics.}
For same-identity reconstruction, we adopt PSNR, $\mathcal{L}_1$ and LPIPS following \citep{tao2024learning} to evaluate the reconstruction quality. We also use FID \citep{heusel2017gans} to measure the realism of the generated video frames. Following \citep{siarohin2019animating}, we employ Average Keypoint Distance (AKD) for motion transfer quality evaluation and Average Euclidean Distance (AED) for identity preservation quality evaluation. 

\vspace{-5pt}
\subsection{Comparison with State-of-the-art Methods}
\vspace{-5pt}

\begin{figure*}[tb]
    
    \begin{center}
  \includegraphics[width=0.99\linewidth]{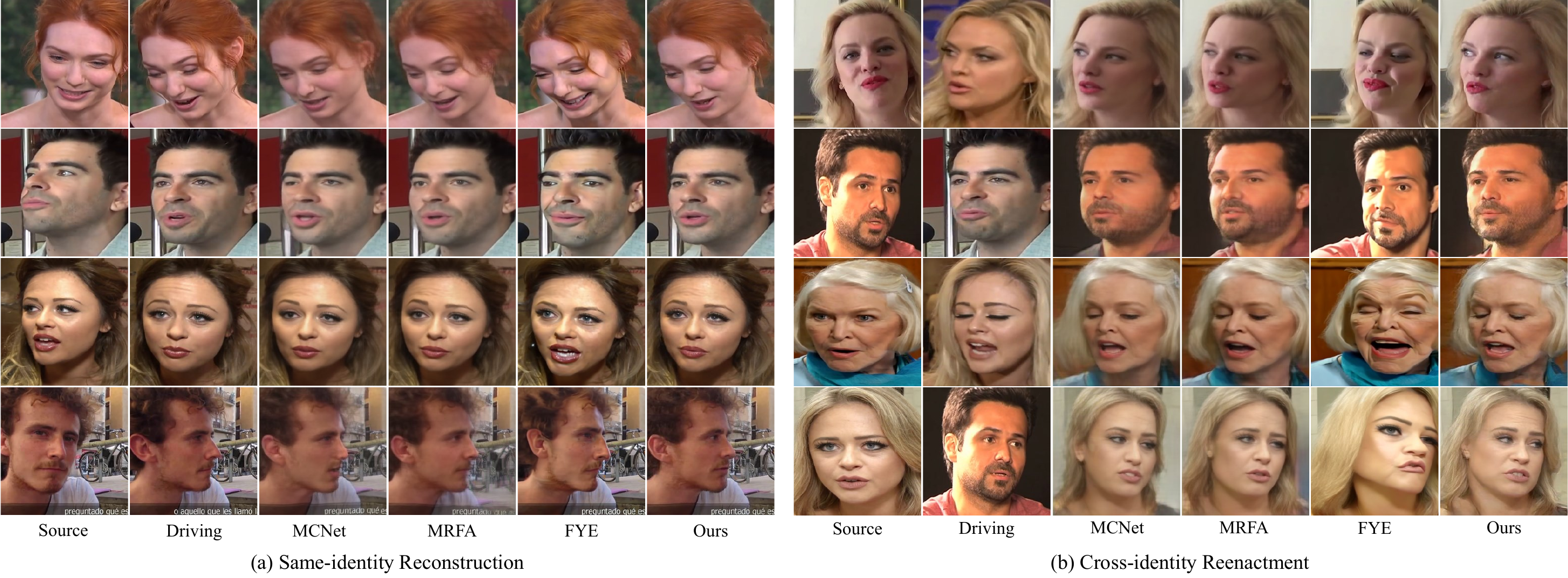}
  \end{center} 
    \vspace{-20pt}
    \caption{Qualitative comparison with state-of-the-art methods for (a) same-identity reconstruction and (b) cross-identity reenactment on the VoxCeleb1 and CelebV-HQ datasets. Our method can generate high-quality facial images \textcolor{black}{even under} large pose variations.}
    \label{fig:sota_comparison}
    %\vspace{-15pt}
\end{figure*}

\begin{table*}[!tb]
  % \vspace{-10pt}
  %\begin{center}
  \centering
  \resizebox{0.92\textwidth}{!}{ %
  \begin{tabular}{l|cccccc|cccccc} %l}
  \toprule
    \multirow{2}{*}{Method}  &  \multicolumn{6}{c|}{VoxCeleb1}  &  \multicolumn{6}{c}{CelebV-HQ} \\
    %\cline{2-7} \cline{8-13}
    
            & FID $\downarrow$ & PSNR $\uparrow$  %& SSIM $\uparrow$ 
             & $\mathcal{L}_1$ $\downarrow$ & LPIPS $\downarrow$ & AKD $\downarrow$ & AED $\downarrow$
             & FID $\downarrow$ & PSNR $\uparrow$  %& SSIM $\uparrow$ 
             & $\mathcal{L}_1$ $\downarrow$ & LPIPS $\downarrow$ & AKD $\downarrow$ & AED $\downarrow$\\
    \midrule
    FOMM \cite{siarohin2019first} 
    & 53.97 & 22.96 %& 0.7463 
    & 0.0474 & 0.2200 & 1.4037 & 0.1509 
    & 78.15 & 20.92 %& 0.6501
    & 0.0685 & 0.2925 & 3.6098 & 0.2955 \\
    %& 15.02 & 23.87 & 0.7757 & 0.0419 & 0.1964 & 1.381 & 0.1304  \\
    %LIA %\cite{wang2021latent} 
    %\\
    \textcolor{black}{LIA \cite{wang2021latent}}
    & 57.51 & 22.88  %& 0.7318 
    & 0.0488 & 0.2293 & 1.5395 & 0.1500
    & 86.64 &  19.58 %& 0.6264
    & 0.0830 & 0.3159  & 3.3632 & 0.3430  \\
    
    DaGAN \cite{hong2022depth} 
    & 51.55 & 22.92 %& 0.7480 
    & 0.0492 & 0.2251 & 1.5740 & 0.1652
    & 99.84 & 20.49 %& 0.6390 
    & 0.0781 & 0.3209 & 7.6075 & 0.3301\\
    %TPSM \cite{Zhao_2022_CVPR}
    %& 53.69 & 24.73 %& 0.7875 
    %& 0.0402 & 0.1974 & 1.2338 & 0.1241
    %& 73.14 & 22.19 %& 0.6752 
    %& 0.0645 & \underline{0.2618} & 4.5840 & 0.2843\\
    %& 13.42 & 25.98 & 0.8228 & 0.0350 & 0.1709 & 1.206 & 0.1080 \\
    MCNet \cite{Hong_2023_ICCV} 
    & 51.45 & 24.59 %& 0.7845 
    & 0.0402 & 0.1996 & 1.2363 & 0.1254
    & 78.33 & 22.20 %& 0.6840
    & 0.0640 & 0.2732 & 4.1386 & 0.2903\\
    %& 13.60 & 26.07 & 0.8213 & 0.0332 & 0.1727 & 1.206 & 0.1068   \\
    MRFA \cite{tao2024learning} 
    & 48.49 & \underline{25.26} %& 0.8007 
    & \underline{0.0370} & \underline{0.1872} & \textbf{1.1823} & \underline{0.1188}  
    & 75.73 & \textbf{22.41} %& 0.6953
    & \underline{0.0625} & \underline{0.2670} & 3.7166 & \textbf{0.2527}
    \\
    %& 12.33 & 26.55 & 0.8328 & 0.0316 & 0.1651 & 1.165 & 0.1021   \\
    AniPortrait~\cite{wei2024aniportrait}
    & 52.65 & 20.15 %& 0.6400
    & 0.0637 & 0.2767 &  2.6543 &0.2623  
    %& 66.67 & 17.37 %& 0.5508
    %& 0.1027 & 0.3263 &  \textbf{1.9444} &0.3517  
    & \textbf{61.56} & 19.71 %& 0.6232
    & 0.0748 & 0.2878 &  \textbf{2.2397} & \underline{0.2739} 
    \\
    
    FYE~\cite{ma2024followyouremoji}
    & \underline{43.25} & 19.54 %& 0.5997
    & 0.0714 & 0.2954 &  2.7071 &  0.2652  
    &\underline{62.55} & 19.58 %&0.6149
    & 0.0802 & 0.3006 & 4.8637 & 0.3029\\
    
    % LivePortrait ~\cite{guo2024liveportrait}
    % & 48.11 & 22.94 %& 0.7333 
    % & 0.0484 & 0.2213 & 1.5516 & 0.1602 
    % &  \textbf{53.88} & 21.25 %& 0.6720
    % & 0.0659 & \textbf{0.2601} & \underline{2.0467} & \underline{0.2718} \\
    
    Ours 
    &  \textbf{43.15}	&\textbf{25.30}	%&0.7916	
    &\textbf{0.0355}	&\textbf{0.1846}	&\underline{1.2039}	&\textbf{0.1071}
    & 71.78	&\underline{22.40}	%&0.6873
    &\textbf{0.0610}	&\textbf{0.2608}	&\underline{3.2562}	&{0.2825}
    \\ 

    \bottomrule
  \end{tabular}
  }%
  \vspace{-8pt}
  \caption{Quantitative comparison with state-of-the-art methods for same-identity reconstruction on VoxCeleb1 and CelebV-HQ dataset\textcolor{black}{s}. Our results are the best on the Voxceleb1 dataset and competitive on the CelebV-HQ dataset.}
  \label{table:sota_sameid}
  %\end{center}
  \vspace{-8pt}
\end{table*}

%We compare our method with a series of open-source state-of-the-art methods including non-diffusion based FOMM \citep{siarohin2019first}, DaGAN \citep{hong2022depth},  TPSM \citep{Zhao_2022_CVPR}, MCNet \citep{Hong_2023_ICCV} and MRFA \citep{tao2024learning}, and diffusion-based AniPortrait \citep{wei2024aniportrait} and Follow-Your-Emoji (FYE) \citep{ma2024followyouremoji}. 

We compare our method with a series of open-source state-of-the-art methods, including non-diffusion-based FOMM~\citep{siarohin2019first}, \textcolor{black}{LIA~\citep{wang2021latent}}, DaGAN \citep{hong2022depth}, MCNet \citep{Hong_2023_ICCV} and MRFA \citep{tao2024learning}, and diffusion-based AniPortrait \citep{wei2024aniportrait} and Follow-Your-Emoji (FYE) \citep{ma2024followyouremoji}. 

%\vspace{-1pt}
\noindent\textbf{Same-identity Reconstruction.}
To evaluate the performance on same-identity reconstruction, we use the first frame of each video as the source image and reconstruct the whole video. Tab.~\ref{table:sota_sameid} presents the quantitative results for same-identity reconstruction. Our method outperforms the other methods almost on all metrics on VoxCeleb1 dataset and remains competitive when generalized to the more challenging CelebV-HQ. Compared with those unsupervised methods~\citep{siarohin2019first,wang2021latent,hong2022depth,Hong_2023_ICCV,tao2024learning}, our method achieves better motion estimation, \emph{e.g.,} our method gets the best AKD score on CelebV-HQ dataset. This result verifies the effectiveness of our designed multi-scale motion codebook and its generalizability. For the results of image quality (\emph{i.e.,} FID, PSNR, $\mathcal{L}_1$, LPIPS), our method outperforms other methods on VoxCeleb1 dataset, even those diffusion methods~\citep{wei2024aniportrait,ma2024followyouremoji} trained with larger-scale datasets. It indicates that our designed multi-scale appearance codebook is capable to compensate for the intermediate warped feature for better talking head video generation.% TODO: Add more analysis
We also show some qualitative results in Fig.~\ref{fig:sota_comparison}a.~Our method can generate plausible unseen facial regions (row 2) and handle large motion (row 4) with accuracy.

\noindent\textbf{Cross-identity Reenactment.} We conduct cross-identity reenactment experiments to validate our method. The qualitative results shown in Fig.~\ref{fig:sota_comparison}b indicate the superiority of our method. Compared to non-diffusion-based MCNet~\citep{Hong_2023_ICCV} and MRFA~\citep{tao2024learning}, our method better preserves facial details (row 1 and 3), even under large motions (row 2 and 4). Diffusion-based FYE~\citep{ma2024followyouremoji} often \textcolor{black}{produces} exaggerated expressions and \textcolor{black}{struggles} to imitate the driving expressions accurately, likely due to its reliance on landmark-based embeddings without explicitly modeling motion. These findings confirm the effectiveness of our framework. More experimental results are shown in the supplementary material.
\subsection{Ablation Study}
\label{subsec:exp-ablation}
\begin{figure}[!tb]
    \vspace{-5pt} 
    %\vspace{-13pt}  
  \begin{center}  \includegraphics[width=0.99\linewidth]{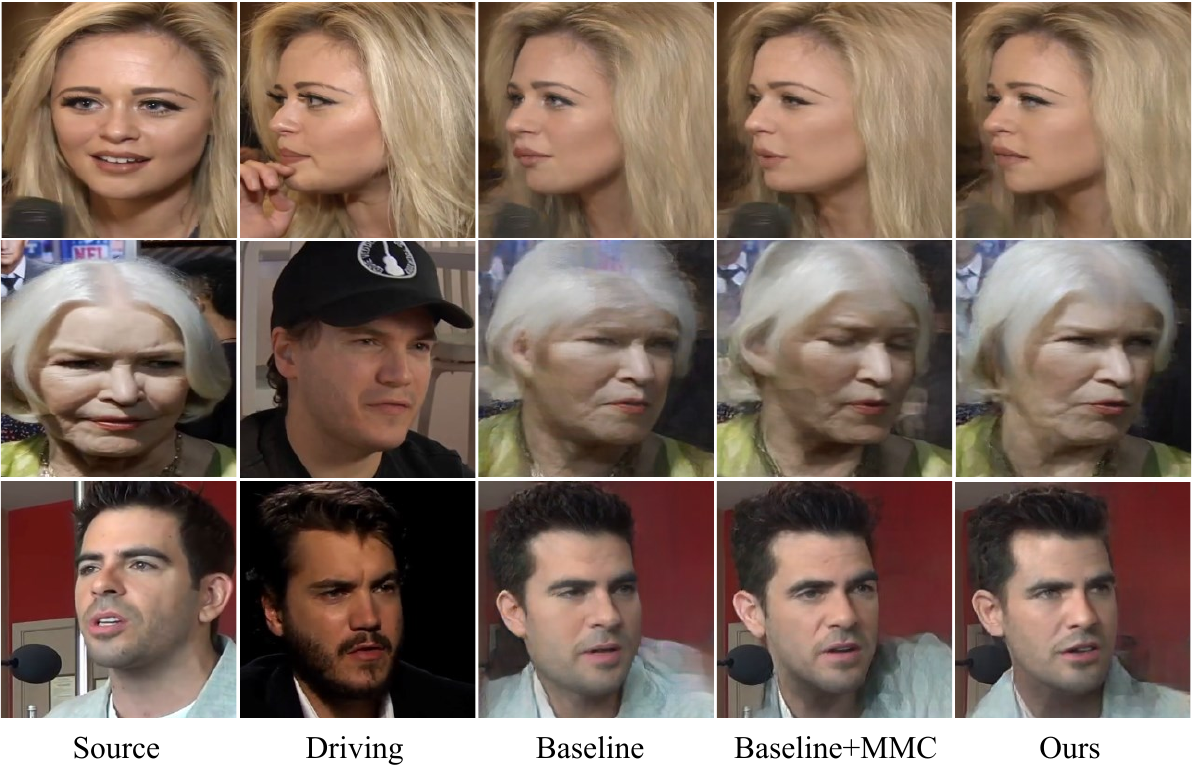}
  \end{center}
    \vspace{-15pt}  
  \caption{Qualitative ablation study on multi-scale motion and appearance codebook compensation. Both motion and appearance codebook compensation contribute to better generation quality.}  \label{fig:ablation_quanlitative}
\vspace{-10pt}  
\end{figure}

% \vspace{-5pt}
We perform ablation studies to assess the effectiveness of the learned multi-scale motion and appearance compensatory codebooks. The model variants in Tab.~\ref{table:ablation_vox1} are as follows: \textbf{(i)} ``Baseline" is the model without any compensatory codebook. \textbf{(ii)} ``Baseline+SMC" includes only a single-scale motion codebook, compensating \textcolor{black}{for} the initial motion flow only at scale $1$, and the compensated result is used to warp multi-scale features. \textbf{(iii)} ``Baseline+MMC" includes a multi-scale motion codebook to compensate for the motion flows across scales. \textbf{(iv)} ``Baseline+MMC+SAC" adds a single-scale appearance codebook to (iii), compensating only the warped feature at scale $1$. \textbf{(v)} ``Baseline+MMC+MAC" is our full model with both multi-scale motion and appearance codebooks. We present the quantitative results in Tab.~\ref{table:ablation_vox1} and qualitative comparisons of (i), (iii), and (v) in Fig.~\ref{fig:ablation_quanlitative}.

\begin{figure}[t]
  \begin{center}  \includegraphics[width=1\linewidth]{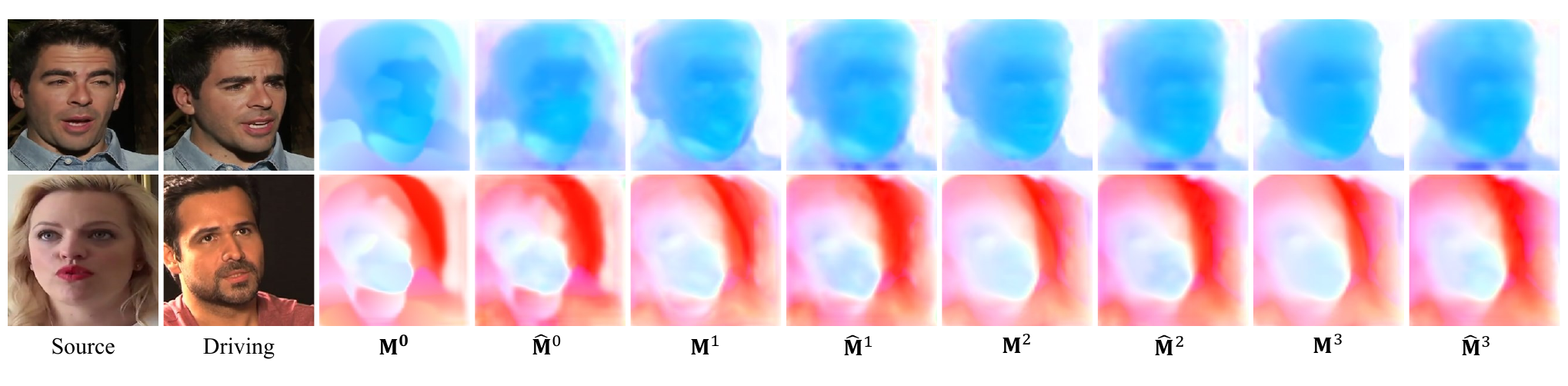}
  \end{center}
  \vspace{-20pt}  
  \caption{Visulization of the original and reconstructed motion flows for different scales. Our multi-scale motion codebook can reconstruct multi-scale motion flows with high quality.}
  \label{fig:effect_motion_cdbk_learning}
  %\vspace{-10pt}
\end{figure}

\begin{figure}[t]
 \vspace{-10pt}  
  \begin{center} \includegraphics[width=1\linewidth]{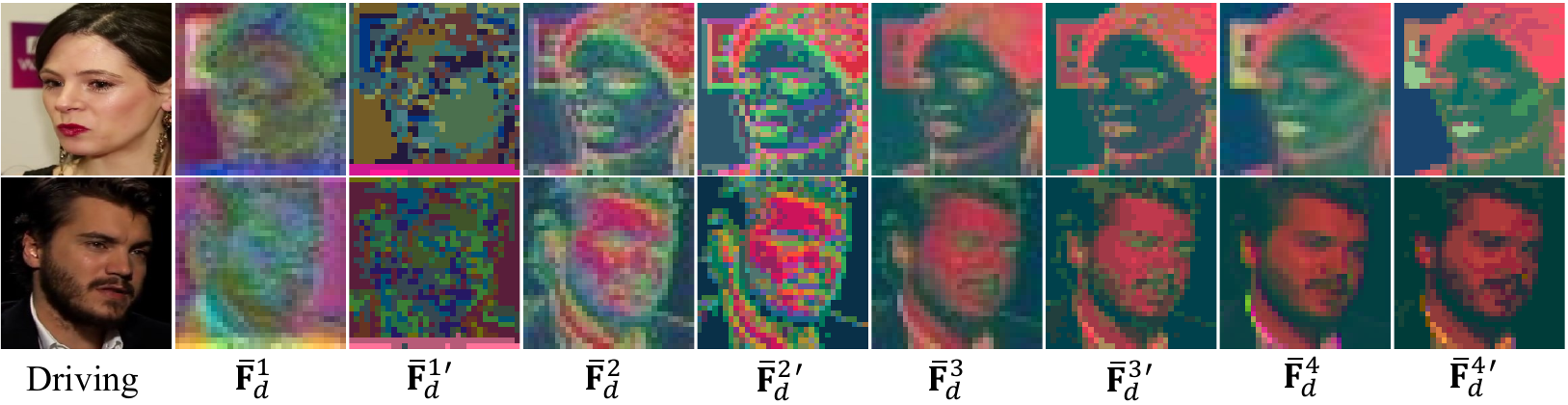}
  \end{center}
   \vspace{-20pt}  
  \caption{Visulization of the original and reconstructed driving features. Our multi-scale appearance codebook can reconstruct multi-scale appearance features with acceptable quantization loss.}
  \label{fig:effect_app_cdbk_learning}
   \vspace{-5pt}  
\end{figure}

\noindent\textbf{Effect of Joint Codebook Learning.}
To evaluate the effectiveness of our jointly learned multi-scale motion and appearance codebooks, we visualize the reconstructed multi-scale motion flows with the motion codebook in Fig.~\ref{fig:effect_motion_cdbk_learning} and appearance features with the appearance codebook in Fig.~\ref{fig:effect_app_cdbk_learning}. For the motion flows, results in Fig.~\ref{fig:effect_motion_cdbk_learning} are output of the motion flow decoder $\mathcal{D}_{M}$ given the quantized motion flow features.
%we use a motion flow decoder $\mathcal{D}_{M}$ to decode the quantized motion flow features and visualize the results. In Fig.~\ref{fig:effect_motion_cdbk_learning}, 
Despite the limited number of codes, our multi-scale motion codebook can reconstruct high-quality motion flows, confirming its ability to capture typical local motion patterns. For appearance features, we visualize the quantized features directly in Fig.~\ref{fig:effect_app_cdbk_learning}. Despite some quantization loss, the multi-scale appearance codebook reconstructs the driving features well with limited codes, \textcolor{black}{demonstrating its} ability to store informative local appearance details. %Note that quantization errors are more noticeable at the feature level, and an image decoder specializing in decoding such quantized features can reduce the perceptual loss at the image level. 
These results demonstrate the effectiveness of joint codebook learning, allowing the codebooks to store expressive local motion and appearance patterns.

\vspace{-5pt}
\begin{table}[tb]
  \centering
  \resizebox{1\linewidth}{!}{
  \begin{tabular}{lllllll} %l}
  \toprule
             & FID $\downarrow$ & PSNR $\uparrow$  %& SSIM $\uparrow$ 
             & $\mathcal{L}_1$ $\downarrow$ & LPIPS $\downarrow$ & AKD $\downarrow$ & AED $\downarrow$ \\
    \midrule
    Baseline 
    & 47.83 & 24.93 
    & 0.0375 & 0.1954 & 1.2384 & 0.1106   \\
    + SMC
    & 49.00 & 24.97  
    & 0.0371 & 0.1917 & 1.2183 & 0.1167   \\
    + MMC
    & 44.86 & 25.27  
    & 0.0360 & 0.1875 & 1.2171 & 0.1076   \\
    + MMC + SAC
    & 44.32 & 25.16 %& 0.7873
    & 0.0362 & 0.1856 & 1.2106 & 0.1091   \\
    + MMC + MAC (Ours) 
    &  \textbf{43.15}	&\textbf{25.30}	%&0.7916	
    &\textbf{0.0355}	&\textbf{0.1846}	&\textbf{1.2039}	&\textbf{0.1071} \\ 
    %&  8.05	&26.65	&0.8239	&0.0300	&0.1629	&1.2005	&0.0951\\ 
    %& 8.56  & 26.52 & 0.8219 & 0.0302 & 0.1647 & 1.195 & 0.0934 \\ % old
    \bottomrule
  \end{tabular}
  }
  \vspace{-10pt}
  \caption{Ablation study on the multi-scale motion and appearance codebook compensation. We present the results for same-identity reconstruction on VoxCeleb1 dataset. 
  }
  \label{table:ablation_vox1}
  \vspace{-20pt}
\end{table}

\begin{figure}[t]
  \begin{center}
  \includegraphics[width=0.99\linewidth]{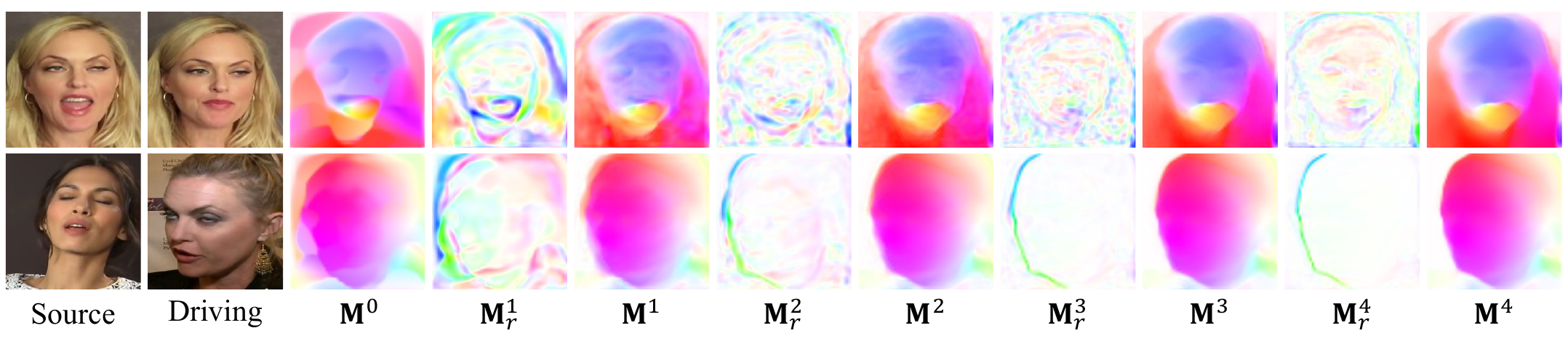}
  \end{center}
  \vspace{-20pt}
  \caption{Visualization of the motion flow compensation process. We present the initial motion flow $\mathbf{M}^0$, the motion flow residual\textcolor{black}{s} $\{\mathbf{M}_r^i\}_{i=1}^N$ and the compensated motion flows $\{\mathbf{M}^i\}_{i=1}^N$.} % for all scales.}
  \label{fig:effect_motion_flow_update}
  \vspace{-10pt}
\end{figure}

\vspace{4pt}
\noindent\textbf{Effect of Multi-scale Motion Codebook Compensation.}
% The second row of Tab.~\ref{table:ablation_vox1} shows that using single-scale motion codebook compensation already improves motion transfer accuracy and image generation quality, with better PSNR, $\mathcal{L}_1$, LPIPS, and AKD compared to the baseline. This highlights the effectiveness of motion codebook compensation for talking head video generation. Multi-scale motion codebook compensation further improves all metrics, emphasizing the importance of handling motion flows at different scales for feature warping. Fig.~\ref{fig:ablation_quanlitative} shows that multi-scale motion compensation better transfers motion (head pose in the first row), reduces artifacts (hair in the second row, face in the third), and preserves source identity (fourth row), demonstrating its effectiveness. We visualize the results in Fig.~\ref{fig:effect_motion_cdbk_learning} and Fig.~\ref{fig:effect_motion_flow_update}. Fig.~\ref{fig:effect_motion_cdbk_learning} compares the compensated motion flow $\mathbf{M}^i$ and quantized motion flow $\hat{\mathbf{M}}^i$, showing that the compensated flow captures more facial detail, validating the transformer-based compensation. Fig.~\ref{fig:effect_motion_flow_update} shows that the initial motion flow $\mathbf{M}^0$ is rough, but our multi-scale compensation refines it iteratively through residuals $\{\mathbf{M}_r^i\}_{i=1}^N$. As the scale increases, the residuals add finer details, resulting in smoother, face-adapted motion flows.
The second row of Tab.~\ref{table:ablation_vox1} shows that using single-scale motion codebook compensation already improves motion transfer and image quality, reflected by better PSNR, $\mathcal{L}_1$, LPIPS, and AKD compared to the baseline. This highlights the effectiveness of motion codebook compensation for talking head video generation. Multi-scale motion codebook compensation further enhances all metrics, underscoring the importance of handling motion flows at different scales for improved feature warping. Fig.~\ref{fig:ablation_quanlitative} shows that multi-scale motion compensation achieves better motion transfer (e.g., head pose in row 1), reduces artifacts (e.g., hair in row 2), and preserves source identity (row 3). Fig.~\ref{fig:effect_motion_flow_update} visualizes the motion flow compensation process, showing that the initial motion flow $\mathbf{M}^0$ is rough, but multi-scale compensation refines it iteratively with residuals $\{\mathbf{M}_r^i\}_{i=1}^N$, adding finer details as the scale increases for smoother, more face-adapted motion flows.

\begin{figure}[t]
  \begin{center} \includegraphics[width=0.99\linewidth]{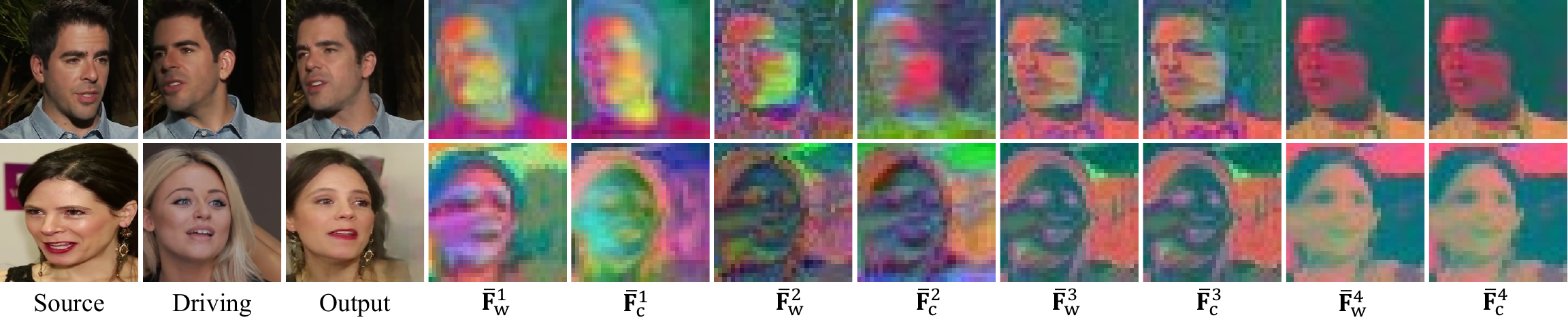} %{figures/results/effect_app_feat_update_originalsize.pdf} %{figures/results/effect_app_feat_update.pdf} %{figures/results/effect_app_feat_update_originalsize.pdf} %{figures/results/effect_app_feat_update.pdf}
  \end{center}
  \vspace{-20pt}
  \caption{Visualization of the appearance compensation results. We present the warped source features \textcolor{black}{$\{\overline{\mathbf{F}}_w^i\}_{i=1}^N$} and the compensated appearance features \textcolor{black}{$\{\overline{\mathbf{F}}_c^i\}_{i=1}^N$}.} % for all scales.}
  \label{fig:effect_app_feat_update}
  \vspace{-10pt}
\end{figure}

\noindent\textbf{Effect of Multi-scale Appearance Codebook Compensation.}
The fourth row in Tab.~\ref{table:ablation_vox1} shows that adding single-scale appearance codebook compensation on top of multi-scale motion codebook compensation improves FID, LPIPS, and AKD, indicating that appearance codebook compensation refines warped source features for more realistic image generation. However, there is a slight drop in PSNR, $\mathcal{L}_1$, and AED, likely due to a conflict between scale $1$ compensated appearance features and other warped features with warping artifacts. The fifth row shows consistent improvement, suggesting \textcolor{black}{that} multi-scale appearance codebook compensation resolves this conflict and further refines warped features across scales, boosting overall performance. The last two columns in Fig.~\ref{fig:ablation_quanlitative} also verify that multi-scale appearance codebook compensation leads to more accurate motion (e.g., the mouth in row 1, eyes in row 2, and shoulders in row 3) with realistic facial details (e.g., the hair in row 2). Additionally, Fig.~\ref{fig:effect_app_feat_update} visualizes the compensated feature maps at different scales, showing more complete facial shapes and details compared to the warped feature \textcolor{black}{{$\overline{\mathbf{F}}_w^i$}}. These results validate the effectiveness of multi-scale appearance codebook compensation.

% The fourth row in Tab. \ref{table:ablation_vox1} shows that adding single-scale appearance codebook compensation based on multi-scale motion codebook compensation improves FID, LPIPS and AKD, which means appearance codebook compensation can refine the warped source features to generate realistic images. There is a slight drop in PSNR, $\mathcal{L}_1$ and AED, which may due to a conflict between the compensated appearance feature at scale 1 and other warped source features containing warping artifacts. The fifth row in Tab. \ref{table:ablation_vox1} shows a consistent improvement over the third and fourth row, which suggests multi-scale appearance codebook compensation can relieve the conflict in single-scale appearance codebook compensation and help refine the warped source features at different scales to boost the overall performance. The last two columns in Fig. \ref{fig:ablation_quanlitative} also demonstrate that multi-scale appearance codebook compensation can help generate images with more accurate motion (the mouth in the first row, the eyes in the second row and the shoulders in the fourth row) and more realistic facial details (the face in the third row). Additionally, we visualize the compensated feature map at different scales in Fig.~\ref{fig:effect_app_feat_update}. The compensated feature maps contain more facial details compared with the warped feature $\mathbf{F}_w^i$. These results verify the effectiveness of multi-scale appearance codebook compensation joint with multi-scale motion codebook compensation.

% flow update process, appearance update process.

\vspace{-8pt}
\section{Conclusion}
 \vspace{-5pt}

In this paper, we have presented a novel framework that jointly learns multi-scale motion and appearance compensatory codebooks to enhance the motion flows and appearance features for talking head video generation. The motion and appearance codebooks store local motion and appearance patterns learned from the entire dataset at different scales, and our multi-scale motion and appearance codebook compensation module retrieves useful codes from the codebooks with a transformer-based strategy at different scales to refine motion flows and appearance features for image generation.~Extensive results verify the effectiveness of our joint motion and appearance codebook learning and compensation, synergizing both~for~high-quality~talking~video~generation.
%\vspace{-15pt}
% \input{sec/X_suppl}

\noindent\textbf{Acknowledgements.}
This research is supported in part by the Early Career Scheme of the Research Grants Council (RGC) of the Hong Kong SAR under grant No. 26202321, SAIL Research Project, HKUST-Zeekr Collaborative Research Fund, HKUST-WeBank Joint Lab Project, and Tencent Rhino-Bird Focused Research Program.

{
    \small
    \bibliographystyle{ieeenat_fullname}
    \bibliography{arxiv}
}

% WARNING: do not forget to delete the supplementary pages from your submission 
\clearpage
\setcounter{page}{1}
\maketitlesupplementary

\section{Additional Implementation Details}
We perform multi-scale compensation across $N=4$ scales. We employ the keypoint-based motion flow estimator from FOMM~\citep{siarohin2019first}. The multi-scale motion flows are estimated at a size of $64\times 64$. We use convolution layers to encode the motion flows into a latent motion flow space of size $32\times 32\times 32$ and set the multi-scale motion codebook size to $K=1024$ and $ d_m=32$. We also use convolution layers to decode the quantized motion flow features while adopting the motion flow updater in MRFA~\citep{tao2024learning} as our motion flow residual decoder. We employ the image encoder and decoder architecture from VQGAN~\citep{esser2021taming} and further encode the multi-scale appearance features into a size of $32\times 32 \times 256$. The multi-scale appearance codebook size is set to $T=1024$ and $ d_a=256$. 

We follow the unsupervised training pipeline from FOMM~\citep{siarohin2019first}, where the source and driving frames are extracted from the same video, and our framework learns to reconstruct the driving frame. 
For the training objective, we use the perceptual loss from FOMM~\citep{siarohin2019first} along with the L1 loss as the image reconstruction loss, and we set the loss weights as $\lambda_{adv}=0.8$, $\lambda^1=0.5$, $\lambda_{recon,m}=32$ and $\beta=0.25$. The entire framework is trained end-to-end utilizing the Adam optimizer, with a learning rate set to $8\times 10^{-5}$ and a batch size of 16 for 250K iterations on four NVIDIA RTX 3090 GPUs.

\section{More Details on Experiments}
\subsection{Additional Details on the Compared Methods}
We evaluate the performance of the compared methods using their released pre-trained models, and we present the training datasets used for each method in Tab.~\ref{table:sota_crossid}. All the GAN-based methods~\citep{siarohin2019first,wang2021latent,hong2022depth,Hong_2023_ICCV,tao2024learning} and our method are trained on the VoxCeleb1~\citep{nagrani2017voxceleb} training set, while the diffusion-based methods AniPortrait~\cite{wei2024aniportrait} and Follow-Your-Emoji (FYE)~\cite{ma2024followyouremoji} are trained on larger-scale datasets, including VFHQ~\citep{xie2022vfhq}, CelebV-HQ~\citep{zhu2022celebv}, HDTF~\citep{Zhang_2021_CVPR}, and their own collected dataset~\cite{ma2024followyouremoji}. 

\subsection{More Experimental Results}

% \subsubsection{Video Demo}
% We present a video demo along with supplementary material, which includes video results from the ablation study and state-of-the-art comparisons to demonstrate the effectiveness of our video generation approach.
\subsubsection{Video Results}
We present video results for the ablation study and state-of-the-art comparisons on the project page\footnote{\href{https://shaelynz.github.io/synergize-motion-appearance/}{https://shaelynz.github.io/synergize-motion-appearance/}} to demonstrate the effectiveness of our video generation approach.

\subsubsection{More Comparison Results}

\begin{figure}[tb]
  %\vspace{-5pt}
  \begin{center}  \includegraphics[width=1\linewidth]{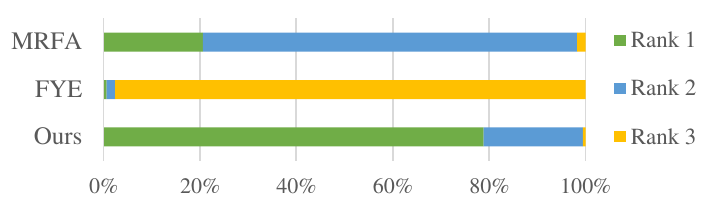}
  \end{center}
  \vspace{-15pt}
  \caption{User study results ranking the quality of videos generated by different methods.} \label{fig:user_study}
  %\vspace{-15pt}
\end{figure}

\noindent\textbf{Cross-identity Reenactment.}
In the absence of ground truth for cross-identity reenactment, we conduct a user study comparing our approach to recent state-of-the-art methods, including a GAN-based model (MRFA~\citep{tao2024learning}) and a diffusion-based model (Follow-You-Emoji (FYE)~\citep{ma2024followyouremoji}). We randomly selected 10 source-driving pairs and asked 30 participants to evaluate the generated videos based on appearance realism, motion naturalness, and overall quality. The results shown in Fig.~\ref{fig:user_study} indicate that users prefer our method, confirming its superiority. 

\begin{table}[!h]
  \centering
  \resizebox{1\linewidth}{!}{
  \begin{tabular}{lcccc} %lll} %l}
  \toprule[1pt]
     Method   & Training Dataset     & FID $\downarrow$ &CSIM $\uparrow$  & ARD $\downarrow$ \\ %& AUH $\downarrow$ \\
    \midrule

    AniPortrait~\cite{wei2024aniportrait} & VFHQ~\citep{xie2022vfhq}, CelebV-HQ~\citep{zhu2022celebv}
    &  \textit{66.61} & \textit{0.7226}
    & \textit{2.9146}
    \\
    FYE~\cite{ma2024followyouremoji} & HDTF~\citep{Zhang_2021_CVPR}, VFHQ~\citep{xie2022vfhq}, their collected dataset~\cite{ma2024followyouremoji}
    & \textit{60.05} & \textit{0.7558} 
    & \textit{3.0822} \\

    \midrule
    
    FOMM \cite{siarohin2019first} &  VoxCeleb1~\citep{nagrani2017voxceleb}
    & 80.00 & 0.6010
    & 1.8331     \\
    
    \textcolor{black}{LIA \cite{wang2021latent}} & VoxCeleb1~\citep{nagrani2017voxceleb}
    & \textbf{72.55}  & \textbf{0.6505}
    & 2.5404  \\
    
    DaGAN \cite{hong2022depth} &  VoxCeleb1~\citep{nagrani2017voxceleb}
    & 85.32 & 0.5743
    & 2.0604     \\
    %TPSM \cite{Zhao_2022_CVPR}
    %& 76.26 &  0.6179 
    %& 1.4985 &    \\
    MCNet \cite{Hong_2023_ICCV} &  VoxCeleb1~\citep{nagrani2017voxceleb}
    & 82.72 & 0.5618
    & 1.6970    \\
    MRFA~\cite{tao2024learning} &  VoxCeleb1~\citep{nagrani2017voxceleb}
    & {77.63}	& {0.5962}
    & \textbf{1.5903} 	 \\ 
    Ours &  VoxCeleb1~\citep{nagrani2017voxceleb}
    & \underline{76.47} & \underline{0.6142}
    & \underline{1.6234} 
    \\
    
    %\midrule
    
    \bottomrule[1pt]
  \end{tabular}
    }
  \caption{Quantitative comparison for cross-identity reenactment on VoxCeleb1 dataset. AniPortrait~\cite{wei2024aniportrait} and Follow-Your-Emoji (FYE)~\cite{ma2024followyouremoji} are trained on much larger-scale datasets and are not suitable for a direct comparison.}
  \label{table:sota_crossid}
  
  %\vspace{-10pt}
\end{table}

\begin{figure*}[tb]
    
    \begin{center}
  \includegraphics[width=0.99\linewidth]{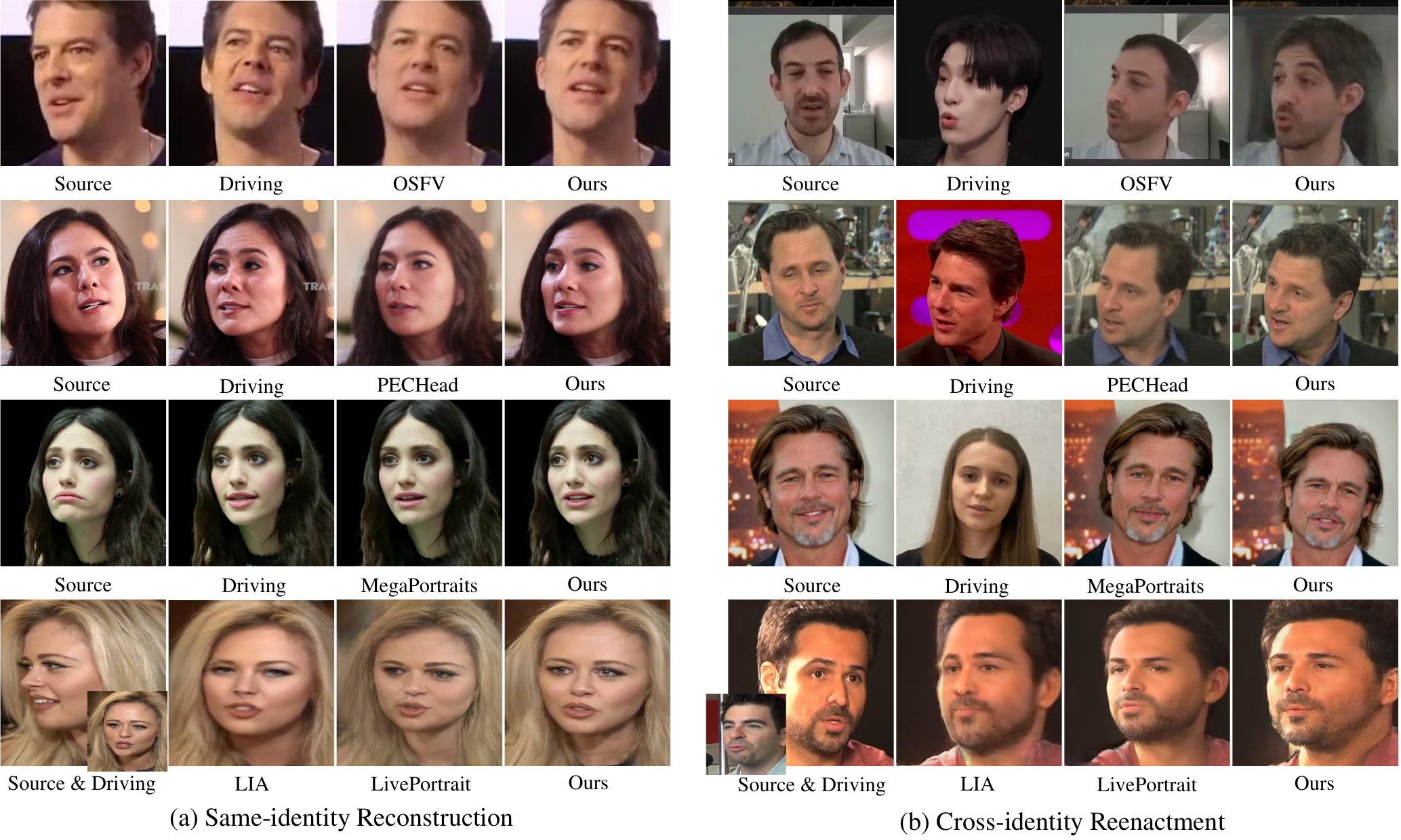}
  \end{center} 
    \vspace{-20pt}
    \caption{\color{black} Qualitative comparison with more state-of-the-art approaches for (a) same-identity reconstruction and (b) cross-identity reenactment on VoxCeleb1 or examples from the corresponding papers or project pages for closed-source methods (\emph{i.e.}, OSFV~\cite{wang2021one}, PECHead~\cite{gao2023high}, and MegaPortraits~\cite{drobyshev2022megaportraits}). Our method better mimics the driving motion and preserves more facial details.}
    \label{supp:more_sota_comparison}
    %\vspace{-15pt}
\end{figure*}

We also present quantitative comparison results for cross-identity reenactment in Tab.~\ref{table:sota_crossid}. We use FID~\cite{heusel2017gans} for image quality evaluation, Average Rotation Distance (ARD) for motion transfer evaluation following~\cite{tao2024learning}, and cosine similarity (CSIM) for identity preservation following~\cite{Ha_Kersner_Kim_Seo_Kim_2020}. Diffusion-based AniPortrait~\cite{wei2024aniportrait} and Follow-Your-Emoji (FYE)~\cite{ma2024followyouremoji} are trained on much larger-scale datasets and are excluded from the comparison. 
%Our method performs the best in terms of image quality and identity preservation, and is comparable in terms of motion transfer, confirming the effectiveness of our approach.
\textcolor{black}{Our method generally demonstrates the highest overall performance, confirming its effectiveness. LIA~\cite{wang2021latent} is slightly better in image quality and identity preservation, as it uses latent codes instead of keypoints as the motion representation, which helps appearance preservation. However, its motion transfer quality is much worse. We also provide a qualitative comparison with LIA in Fig.~\ref{supp:more_sota_comparison} where our method mimics the driving motion better.}
Although MRFA~\cite{tao2024learning} can transfer motion well, its output frame quality may be low and it may not preserve source identity effectively. 
Although diffusion-based methods~\cite{wei2024aniportrait,ma2024followyouremoji} can achieve even better image quality and identity preservation performance, they struggle to transfer motion faithfully, which leads to undesirable visual quality. 

{\color{black}
\noindent\textbf{Comparison with more state-of-the-art approaches.}
We additionally provide comparisons with more state-of-the-art approaches, including an open-source method (\emph{i.e.}, LivePortrait~\cite{guo2024liveportrait}) and several closed-source methods (\emph{i.e.}, OSFV~\cite{wang2021one}, PECHead~\cite{gao2023high}, and MegaPortraits~\cite{drobyshev2022megaportraits}). The qualitative comparison in Fig.~\ref{supp:more_sota_comparison} highlights the advantages of our method in the preservation of facial details and expression transfer. We also provide a quantitative comparison with the open-source LivePortrait~\cite{guo2024liveportrait} in Tab.~\ref{supp:table:sota_sameid}. Although LivePortrait is trained on significantly larger datasets and is unsuitable for a direct fair comparison, our method can still outperform it on all metrics for same-identity reconstruction.

\begin{table}[!ht]
  % \vspace{-10pt}
  %\begin{center}
  \centering
  \resizebox{1\linewidth}{!}{ %
  \begin{tabular}{l|c|cccccc|ccc}%|cccccc} %l}
  \toprule[1pt]
    \multirow{2}{*}{{\large Method}}  & { {\large \# Training}} & \multicolumn{6}{c|}{ {\large Same-identity Reconstruction}} & \multicolumn{3}{c}{ {\large Cross-identity Reenactment}}
    %&  \multicolumn{6}{c}{CelebV-HQ} 
    \\
    %\cline{2-7} \cline{8-13}
    
          & {\large Video Frames}  & {\large FID} $\downarrow$ & {\large PSNR} $\uparrow$  %& SSIM $\uparrow$ 
             &{\large  $\mathcal{L}_1$} $\downarrow$ &{\large  LPIPS} $\downarrow$ &{\large  AKD} $\downarrow$ & {\large AED} $\downarrow$ &  {\large FID} $\downarrow$ &  {\large CSIM} $\uparrow$  & {\large ARD} $\downarrow$
             %& FID $\downarrow$ & PSNR $\uparrow$  %& SSIM $\uparrow$ 
             %& $\mathcal{L}_1$ $\downarrow$ & LPIPS $\downarrow$ & AKD $\downarrow$ & AED $\downarrow$
             \\
    \midrule
    %\multirow{2}{*}
    {{\large LivePortrait~\cite{guo2024liveportrait}}} %\cite{guo2024liveportrait}
    & %{\footnotesize VoxCeleb, MEAD, RAVDESS, AAHQ,} 
    {\large 69M}
    & %\multirow{2}{*}
    { {\large{\textit{48.11}}}} & %\multirow{2}{*}
    { {\large{\textit{22.94}}}} 
    & %\multirow{2}{*}
    { {\large{\textit{0.0484}}}} & %\multirow{2}{*}
    { {\large{\textit{0.2213}}}} & %multirow{2}{*}
    {  {\large \textit{1.5516}}} &  %\multirow{2}{*}
    { 
    {\large \textit{0.1602}}} 
    & %\multirow{2}{*}
    { {\large\textbf{\textit{75.95}}}} & %\multirow{2}{*}
    { {\large\textbf{\textit{0.7260}}}}
    & %\multirow{2}{*}
    { {\large\textbf{\textit{1.3497}}}}

    \\
    %& {\footnotesize LightStage, their collected dataset} &&&&&&

    %& \textbf{53.88} &  \underline{21.25} %&  
    %& \underline{0.0659}  & \textbf{0.2601}  & \textbf{2.0467}  & \textbf{0.2718} 
    %\\  

    \midrule
    
     {\large Ours} 
    %& VoxCeleb1
    & {\large 4.3M}
    &   {\large\textbf{43.15}}	& {\large\textbf{25.30}}	%&0.7916	
    & {\large\textbf{0.0355}}	& {\large\textbf{0.1846}}	& {\large\textbf{1.2039}}	& {\large\textbf{0.1071}}

    &  {\large 76.47} &  {\large 0.6142}
    &  {\large {1.6234}} 
    
    %& \underline{71.78}	&\textbf{22.40}	%&0.6873
    %&\textbf{0.0610}	&\underline{0.2608}	&\underline{3.2562}	&\underline{0.2825}
    \\ 

    \bottomrule[1pt]
  \end{tabular}
  }%
  %\vspace{-10pt}
  \caption{\color{black} Quantitative comparison with LivePortrait~\cite{guo2024liveportrait} on VoxCeleb1. LivePortrait, being trained on \emph{significantly larger} data, is unsuitable for a direct comparison.}% and CelebV-HQ dataset.}
  \label{supp:table:sota_sameid}
  %\end{center}
  %\vspace{-13pt}
\end{table}

\noindent\textbf{Inference speed.}
We evaluate the inference speed using an NVIDIA RTX 3090 and provide the results in Tab.~\ref{reb:table:sota_speed}. Our approach shows clear advantages upon recent state-of-the-art methods~\cite{tao2024learning,wei2024aniportrait,ma2024followyouremoji,guo2024liveportrait}, %the diffusion-based AniPortrait and FYE, as well as the non-diffusion-based MRFA and LivePortrait,
indicating its potential for real-time performance.
%Our method is faster than diffusion-based AniPortrait, FYE, and non-diffusion-based MRFA and LivePortrait. This highlights our method's potential to run in real time. 

%\vspace{-10pt}
\begin{table}[tb]
  \begin{center}
  \resizebox{1\linewidth}{!}{
  \begin{tabular}{l|ccccccccc}
  \toprule[1pt]
             %& FOMM & LIA &DaGAN & MCNet 
             & MRFA~\cite{tao2024learning} & AniPortrait~\cite{wei2024aniportrait} & FYE~\cite{ma2024followyouremoji} & LivePortrait~\cite{guo2024liveportrait} & Ours \\
    \midrule
    %Memory (M) $\downarrow$ & & & & & & 9525.3125 & 15163.3125 & 21541.3125 & 6413.3125 \\
    %Output resolution &
    %256 $\times$ 256 & 256$\times$ 256 & 512\times 512 & 512\times 512  & 512\times 512 & 256\times 256
    %\\
    FLOPs  $\downarrow$
    %& 112.76G & \textbf{92.57G} %&  
    %& 169.91G  & 398.60G
    & 403.05G  & 9.18T & 15.04T& 1.31T &  \textbf{352.91G}  \\
    FPS $\uparrow$
    %& \textbf{24.22} & 21.87  
    %& 18.54 &  \textbf{23.39} 
    & 12.41 & 0.36 & 0.39 & 11.28 & \textbf{15.13}  \\ 

    \bottomrule[1pt]
  \end{tabular}
  }
  \end{center}
  \vspace{-10pt}
  \caption{\color{black} Inference speed comparison.
  }
  %\vspace{-13pt}
  \label{reb:table:sota_speed}
\end{table}

}

\subsubsection{Additional Ablation Study}

\noindent\textbf{Effect of the code allocation scheme.}
We propose a novel code allocation scheme for motion and appearance codebooks that assigns different codes to corresponding scales. This allows certain codes to be shared across multiple scales, facilitating the transfer of information between them. To assess the effect of our code allocation scheme, we conduct an ablation study and present results in Tab.~\ref{table:ablation_code_struction}. We compare with two alternative codebook splitting schemes: sharing all codes across all scales and splitting the codes equally among the scales. As demonstrated in Tab.~\ref{table:ablation_code_struction}, our code allocation scheme generally achieves the best overall performance, confirming the superior performance of our code allocation scheme. 

\begin{table}[tb]
  \begin{center}
  \resizebox{1\linewidth}{!}{
  \begin{tabular}{lllllll} %l}
  \toprule[1pt]
          Method   & FID $\downarrow$ & PSNR $\uparrow$  %& SSIM $\uparrow$ 
             & $\mathcal{L}_1$ $\downarrow$ & LPIPS $\downarrow$ & AKD $\downarrow$ & AED $\downarrow$ \\
    \midrule
    Sharing all codes 
    & 43.23 & 25.12 %& 0.7881 
    & 0.0359 & 0.1860 & 1.2124 & \textbf{0.1065}   \\
    Splitting the codes equally
    & \textbf{42.52} & 25.20 %&0.7868
    & 0.0358 & 0.1857 & \textbf{1.1893} & 0.1075  \\
    
    Code Allocation (Ours) 
    &  43.15	&\textbf{25.30}	%&0.7916	
    &\textbf{0.0355}	&\textbf{0.1846}	&1.2039	&0.1071 \\ 

    \bottomrule[1pt]
  \end{tabular}
  }
  \end{center}
  \vspace{-10pt}
  \caption{Ablation study on the code allocation scheme.
  }
  \label{table:ablation_code_struction}
\end{table}

{\color{black}
\noindent\textbf{Effect of the model design.} 
To verify the source of our performance improvement, we compare with a new ``Baseline*", which has parameters comparable to our full model, achieved by increasing the ResBlock channel numbers of our image encoder and decoder. We present the results in Tab.~\ref{reb:table:ablation_design}. Our method significantly improves upon ``Baseline*". The clear performance gap further confirms that the improvement comes from our model design rather than the increased parameters, indicating the effectiveness of our model design.
%We present results in Tab.~\ref{reb:table:ablation_design} where ``Baseline*" is our baseline model with comparable parameters to our full model, achieved by increasing the ResBlock channel numbers of our image encoder and decoder. Our method shows significant improvement, further indicating that the performance improvement comes from the model design.
%which demonstrates the performance improvement from the model design.

%\vspace{-10pt}
\begin{table}[tb]
  \begin{center}
  %\vspace{-10pt}
  \resizebox{1\linewidth}{!}{
  \begin{tabular}{l|c|cccccc} %l}
  \toprule[1pt]
          Method & \# Params (M)  & FID $\downarrow$ & PSNR $\uparrow$  %& SSIM $\uparrow$ 
             & $\mathcal{L}_1$ $\downarrow$ & LPIPS $\downarrow$ & AKD $\downarrow$ & AED $\downarrow$  \\
    \midrule
    Baseline* & 82.2
    & 48.09 & 21.64 %& 0.6863
    & 0.0549 & 0.2480 
    & 2.6798 & 0.2214
    \\
    
    Ours & 82.2
    &  \textbf{43.15}	&\textbf{25.30}	%&0.7916	
    &\textbf{0.0355} &\textbf{0.1846}	& \textbf{1.2039}	& \textbf{0.1071} \\ 

    \bottomrule[1pt]
  \end{tabular}
  }
  \end{center}
  \vspace{-10pt}
  \caption{\color{black} Ablation study on the model design.
  }
  %\vspace{-15pt}
  \label{reb:table:ablation_design}
\end{table}

\noindent\textbf{Codebook size.}
To assess how the codebook size affects the generation speed and quality, we vary the number of codes in the codebooks to achieve different codebook sizes and present results in Tab.~\ref{reb:table:ablation_code_num}. 
Larger codebooks generally improve generation quality by providing sufficient capacity to learn diverse motion and appearance codes, with only a slight decrease in speed/memory performance.
%Generally, generation quality improves with larger codebooks, as they provide sufficient capacity to learn expressive motion and appearance codes, with only a slight decrease in speed/memory usage.
%The generation quality generally increases with the increase of codebook size, for it can provide enough space to learn expressive motion and appearance codes with only a slight decrease in speed/memory usage. 
A small codebook of 256 also performs well, likely because codes are retrieved more frequently during training, allowing for better optimization within the same training iterations. However, its image quality remains limited.
%Meanwhile, the model with a small code number of 256 also achieves good results, which is probably because the codes are retrieved more frequently during training, allowing more sufficient optimization of the codebook given the same training iterations. However, its generated image quality is limited.
%A small code number of 256 can effectively learn to compensate for motion and appearance (AKD and AED in row 1), but the generated image quality is limited with relatively low FID, PSNR, $L_1$ and LPIPS. Although a relatively larger codebook generally achieves inferior results (row 2), with a code number of 1024, the model produces the best overall results, providing enough space to learn expressive motion and appearance codes with only a slight decrease in speed/memory usage.
%\textcolor{orange}{Continue...} 

%\vspace{-10pt}
\begin{table}[tb]
  \begin{center}
  \resizebox{1\linewidth}{!}{
  \begin{tabular}{l|cccccc|ccc}
  \toprule[1pt]
          Number of Codes %& Appearance Code Dimension & Motion Code Dimension  
          & FID $\downarrow$ & PSNR $\uparrow$  %& SSIM $\uparrow$ 
             & $\mathcal{L}_1$ $\downarrow$ & LPIPS $\downarrow$ & AKD $\downarrow$ & AED $\downarrow$ & FLOPs (G) $\downarrow$ & FPS $\uparrow$ & Memory (M) $\downarrow$ \\
    \midrule
    256 %& \multirow{3}{*}{256} & \multirow{3}{*}{32}
    & 47.50  & 25.18 %&  
    & 0.0358 & 0.1861 &  \textbf{1.1970} &  \textbf{0.1039} & \textbf{351.57} & \textbf{15.60} & \textbf{6411} \\
    512 %& &
    & 46.62  & 25.11  %& 0.7855
    & 0.0362 & 0.1877 & 1.2190 & 0.1072 & 352.01 & 15.47 & \textbf{6411} \\
    
    1024 (Ours) %& &
    &  \textbf{43.15}	&\textbf{25.30}	%&0.7916	
    &\textbf{0.0355}	&\textbf{0.1846}	& 1.2039	&0.1071 & 352.91 & 15.13 %10.66 
    & 6413 \\ 

    \bottomrule[1pt]
  \end{tabular}
  }
  \end{center}
  \vspace{-10pt}
  \caption{\color{black} Ablation study on the codebook size. We present the results of different code numbers. }
  \label{reb:table:ablation_code_num}
\end{table}

}

%For both the motion and appearance codebooks, we propose a novel code allocation scheme that assigns different codes to corresponding scales. This allows certain codes to be shared across multiple scales, facilitating the transfer of information between them. To assess the effectiveness of our codebook structure, we conduct an ablation study, as shown in Tab.~\ref{table:ablation_code_struction}. We compare two alternative strategies: sharing all codes across all scales, and splitting the codes equally among the scales. As demonstrated in Tab.~\ref{table:ablation_code_struction}, our code allocation strategy achieves the best overall performance considering all the metrics. These results highlight the effectiveness and superiority of our proposed codebook construction.

\section{Limitation}
A limitation of our method is the appearance leakage problem in cross-identity reenactment, where the face in the generated video tends to have a shape similar to that of the driving face rather than the source face.  This issue arises from the keypoint-based motion flow estimator that we adopt to produce the initial coarse motion flow and the driving keypoints for multi-scale motion codebook compensation. Although this motion flow estimator is robust to non-facial motion, such as hair and neck movement, by learning unsupervised keypoints on talking heads, the keypoints also inherently model facial shapes, which leads to the entanglement of motion and shape. \textcolor{black}{Thus, appearance leakage is a common issue for keypoint-based methods. Our method can effectively alleviate this issue by demonstrating better appearance preservation than other state-of-the-art keypoint-based approaches. As evidenced in Tab.~\ref{table:sota_crossid}, we achieve the highest CSIM score among these approaches, excluding LIA~\cite{wang2021latent}, which uses latent codes instead of keypoints for motion representation. This issue can also be mitigated through relative motion transfer~\cite{siarohin2019first}, which} is widely adopted by previous methods~\cite{hong2022depth,Hong_2023_ICCV,tao2024learning}. 
%Relative motion transfer \cite{siarohin2019first} can alleviate this problem and is widely adopted by previous methods~\cite{hong2022depth,Hong_2023_ICCV,tao2024learning}.

\end{document}